\documentclass[10pt,twocolumn,letterpaper]{article}

\usepackage[pagenumbers]{cvpr}  
\usepackage{tabularx}
\usepackage{multirow}

\usepackage[per-mode=symbol, group-digits=integer, group-minimum-digits = 4, detect-all=true, input-symbols={()}, retain-explicit-plus]{siunitx}

\usepackage[percent]{overpic} 
\usepackage{fp} 
\usepackage{graphicx}
\usepackage{tikz}
\usetikzlibrary{positioning}
\usetikzlibrary{shapes.geometric, shapes, arrows, arrows.meta, bending}
\usetikzlibrary{backgrounds}
\usetikzlibrary{spy}
\usepackage{pgfplots}
\usepackage{bbding}
\usepackage{dsfont}
\usepackage{fontawesome5}
\usepackage{calc}
\pgfplotsset{compat=1.17}
\usepackage[outline]{contour} 
\contourlength{0.02em}
\usepackage[table]{xcolor}
\usepackage{makecell}
\usepackage{adjustbox}
\usepackage{pifont}
\usepackage[scaled=0.97]{newtxtt}
\usepackage{utfsym}
\usepackage[super]{nth}
\newcommand\ours{INSID3}

\newcommand*{\inparagraph}[1]{\smallskip\noindent\textbf{#1}\hspace{0.4em}}

\DeclareMathOperator*{\argmax}{arg\,max}

\definecolor{cvprcolor}{RGB}{127,127,255}
\definecolor{softrow}{RGB}{245,245,245}
\definecolor{indomain}{gray}{0.55}
\newcommand{\ind}[1]{\textcolor{indomain}{#1}}

\definecolor{tud0d}{RGB}{83,83,83}
\definecolor{tud0c}{RGB}{137,137,137}
\definecolor{tud0b}{RGB}{181,181,181}
\definecolor{tud0a}{RGB}{220,220,220}
\definecolor{tud1a}{RGB}{93,133,195}
\definecolor{tud2a}{RGB}{0,156,218}
\definecolor{tud3a}{RGB}{80,182,149}
\definecolor{tud4a}{RGB}{175,204,80}
\definecolor{tud5a}{RGB}{221,223,72}
\definecolor{tud6a}{RGB}{255,224,92}
\definecolor{tud7a}{RGB}{248,186,60}
\definecolor{tud8a}{RGB}{238,122,52}
\definecolor{tud9a}{RGB}{233,80,62}
\definecolor{tud10a}{RGB}{201,48,142}
\definecolor{tud11a}{RGB}{128,69,151}
\definecolor{tud1b}{RGB}{0,90,169}
\definecolor{tud2b}{RGB}{0,131,204}
\definecolor{tud3b}{RGB}{0,157,129}
\definecolor{tud4b}{RGB}{153,192,0}
\definecolor{tud5b}{RGB}{201,212,0}
\definecolor{tud6b}{RGB}{253,202,0}
\definecolor{tud7b}{RGB}{245,163,0}
\definecolor{tud8b}{RGB}{236,101,0}
\definecolor{tud9b}{RGB}{230,0,26}
\definecolor{tud10b}{RGB}{166,0,132}
\definecolor{tud11b}{RGB}{114,16,133}
\definecolor{tud1c}{RGB}{0,78,138}
\definecolor{tud2c}{RGB}{0,104,157}
\definecolor{tud3c}{RGB}{0,136,119}
\definecolor{tud4c}{RGB}{127,171,22}
\definecolor{tud5c}{RGB}{177,189,0}
\definecolor{tud6c}{RGB}{215,172,0}
\definecolor{tud7c}{RGB}{210,135,0}
\definecolor{tud8c}{RGB}{204,76,3}
\definecolor{tud9c}{RGB}{185,15,34}
\definecolor{tud10c}{RGB}{149,17,105}
\definecolor{tud11c}{RGB}{97,28,115}
\definecolor{tud1d}{RGB}{36,53,114}
\definecolor{tud2d}{RGB}{0,78,115}
\definecolor{tud3d}{RGB}{0,113,94}
\definecolor{tud4d}{RGB}{106,139,55}
\definecolor{tud5d}{RGB}{153,166,4}
\definecolor{tud6d}{RGB}{174,142,0}
\definecolor{tud7d}{RGB}{190,111,0}
\definecolor{tud8d}{RGB}{169,73,19}
\definecolor{tud9d}{RGB}{156,28,38}
\definecolor{tud10d}{RGB}{115,32,84}
\definecolor{tud11d}{RGB}{76,34,106}

\definecolor{semantic}{RGB}{37, 150, 190}
\definecolor{part}{RGB}{102, 51, 1}
\definecolor{personalized}{RGB}{0, 0, 101}

\definecolor{myorange}{RGB}{255, 127, 15}
\definecolor{myblue}{RGB}{52, 153, 255}
\definecolor{mypurple}{RGB}{178, 102, 255}

\newcommand\pckTen{$\text{PCK@}0.10$}

\newlength{\plotW}
\newlength{\plotH}

\newcolumntype{C}[1]{>{\centering\arraybackslash}p{#1}}
\newcommand{\spillC}[2]{%
  \makebox[0pt]{\hspace*{\dimexpr#1/2\relax}\makebox[0pt][c]{#2}}%
  \rule{#1}{0pt}%
}

\definecolor{cvprblue}{rgb}{0.21,0.49,0.74}
\usepackage[pagebackref,breaklinks,colorlinks,allcolors=cvprblue]{hyperref}

\usepackage{pifont}
\usepackage[table]{xcolor}
\usepackage{subcaption} 
\usepackage{tikz}
\usetikzlibrary{calc}

\usepackage{booktabs}   
\usepackage{array}      
\usepackage{makecell}   
\usepackage{multirow}
\usepackage{graphicx} 
\usepackage{textcomp} 
\usepackage{stfloats}
\usepackage{colortbl}

\title{INSID3: Training-Free \underline{I}\underline{n}-Context \underline{S}egmentation w\underline{i}th \underline{D}INOv\underline{3}}

\newcommand{\myparagraph}[1]{\smallskip\noindent\textbf{#1}\hspace{0.4em}}
\newcommand{\myparagraphnospace}[1]{\smallskip\noindent\textbf{#1}}

\newcommand{\authorstep}{\hspace{0.75cm}}
\newcommand{\affiliationstep}{\hspace{0.75cm}}

\author{
Claudia Cuttano\textsuperscript{\normalfont{}* 1,2}
\authorstep Gabriele Trivigno\textsuperscript{\normalfont{}* 1}
\authorstep Christoph Reich\textsuperscript{\normalfont{}\,2,3,5,6}\\
Daniel Cremers\textsuperscript{\normalfont{}\,3,5,6}
\authorstep Carlo Masone\textsuperscript{\normalfont{}\,1}
\authorstep Stefan Roth\textsuperscript{\normalfont{}\,2,4,5}\\[0pt]
\small{\textsuperscript{1}Politecnico di Torino\affiliationstep \textsuperscript{2}TU Darmstadt\affiliationstep \textsuperscript{3}TU Munich\affiliationstep \textsuperscript{4}hessian.AI\affiliationstep \textsuperscript{5}ELIZA\affiliationstep \textsuperscript{6}MCML\affiliationstep
\textsuperscript{*}equal contribution}\\[-2pt]\small {\url{https://visinf.github.io/INSID3}}}

\usepackage{fancyhdr}
\usepackage{setspace}

\fancypagestyle{firststyle}
{

	\fancyhf{}
	\lfoot{{\footnotesize\begin{spacing}{.5}\parbox{\linewidth}{\vspace{-0.5em}%
					To appear in \emph{Proceedings of the IEEE/CVF Conference on Computer Vision and Pattern Recognition}, Denver, CO, USA, 2026.%
					\\\hrule\vspace{\baselineskip}
					\copyright~2026 IEEE. Personal use of this material is permitted. Permission from IEEE must be obtained for all other uses, in any current or future media, including reprinting/republishing this material for advertising or promotional purposes, creating new collective works, for resale or redistribution to servers or lists, or reuse of any copyrighted component of this work in other works. 
	}\end{spacing}}}
}

\begin{document}
\twocolumn[{%
\renewcommand\twocolumn[1][]{#1}%
\maketitle
\vspace{-1.195em}
\includegraphics[width=0.996\textwidth]{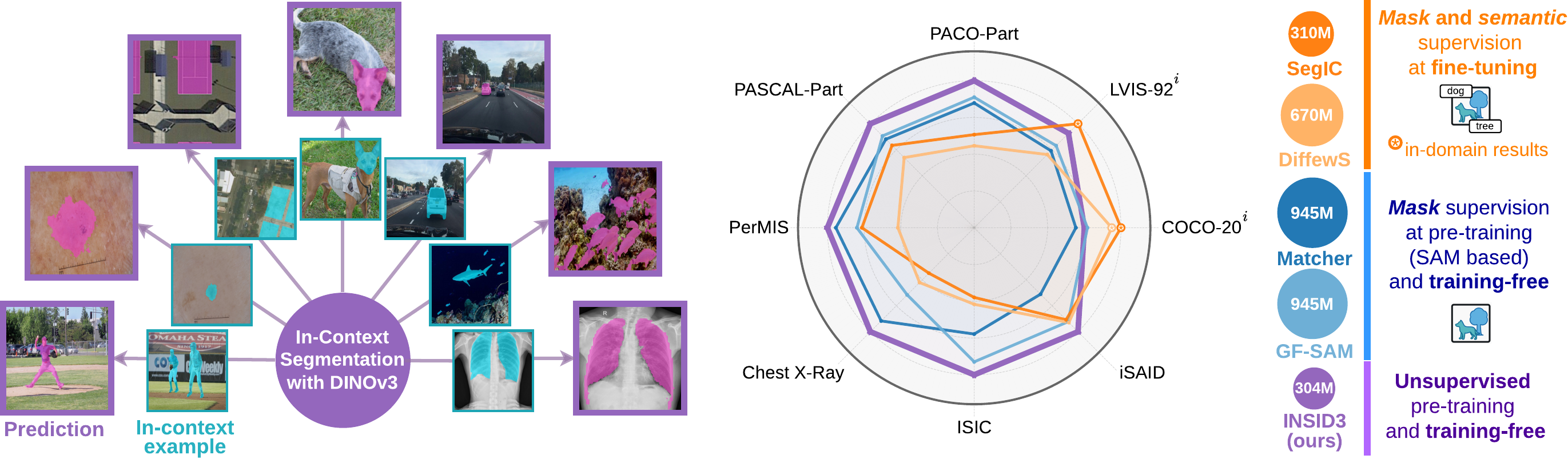}
\vspace{-0.65em}
\captionof{figure}{\textbf{Results and overview of \ours{}, our training-free in-context segmentation approach}. \ours{} performs {in-context segmentation} directly from {DINOv3~\cite{Simeoni:2025:Dinov3}} features, without any decoder, fine-tuning, or model composition.
\emph{(left)} A single annotated example guides the model to segment any concept, {from object parts to medical images and aerial views}. \emph{(right)} Comparing generalization across datasets and segmentation granularities: fine-tuned methods \emph{(\textcolor{myorange}{orange})} excel in-domain \emph{(\includegraphics[height=7.5pt]{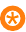})} but degrade out of distribution, while SAM-based pipelines \emph{(\textcolor{myblue}{blue})} generalize better but rely on large, multi-stage architectures. \ours{} \emph{(\textcolor{mypurple}{purple})} achieves the strongest generalization with a single backbone, revealing that robust segmentation can emerge directly from the dense self-supervised representations of DINOv3.}
\label{fig:teaser}
\vspace{1.5\baselineskip}
}]

{\centering\large \textbf{Abstract}\par}

\noindent\emph{In-context segmentation (ICS) aims to segment arbitrary concepts, \eg, objects, parts, or personalized instances, given one annotated visual examples. Existing work relies on \emph{(i)} fine-tuning vision foundation models (VFMs), which improves in-domain results but harms generalization, or \emph{(ii)} combines multiple frozen VFMs, which preserves generalization but yields architectural complexity and fixed segmentation granularities. We revisit ICS from a minimalist perspective and ask: Can a single self-supervised backbone support both semantic matching and segmentation, without any supervision or auxiliary models? 
We show that scaled-up dense self-supervised features from DINOv3 exhibit strong spatial structure and semantic correspondence. We introduce \ours{}, a training-free approach that \emph{segments concepts} at varying granularities only from frozen DINOv3 features, given an in-context example.
\ours{} achieves state-of-the-art results across one-shot semantic, part, and personalized segmentation, outperforming previous work by \SI{+7.5}{\%} mIoU, while using 3$\times$ fewer parameters and without any mask or category-level supervision.
}
\thispagestyle{firststyle}
\section{Introduction\label{sec:introduction}}
Understanding visual scenes is a fundamental task with applications in autonomous driving~\cite{Janai:2020:AVS, Cordts:2016:TCD}, robotics~\cite{Geiger:2013:KIT}, augmented reality~\cite{Ko:2020:AR}, or medical image analysis~\cite{Wang:2022:MED}. In-context segmentation (ICS)~\cite{Liu:2023:Matcher, Meng:2024:SEGiC, Zhang:2023:PerSAM} approaches the task of segmenting arbitrary concepts, such as \textit{objects}, \textit{parts}, or \textit{personalized instances} in images, given one or more annotated examples at inference time, \cf \cref{fig:teaser} \emph{(left)}. This holistic and open-world scene understanding task requires adaptability to different reference annotations and domains, sharing the spirit of adapting large language models (LLMs) through contextual instructions to novel tasks~\cite{Brown:2020:GPT, Ouyang:2022:InstructGPT,Chowdhery:2023:Palm, Touvron:2023:Llama}.

ICS requires reliable visual correspondences between annotated reference examples and target images. Previous work showed that such visual correspondences emerge in features of vision foundation models (VFMs) \cite{Zhang:2023:Tale, Tang:2023:Dift}.
Based on this, recent work has explored how to endow VFMs with explicit segmentation capabilities. For instance, \cite{Meng:2024:SEGiC, Liu:2024:SINE, Zhu:2024:Unleashing} augment a frozen DINOv2 \cite{Oquab:2023:Dinov2} by training a segmentation decoder on top or fine-tune a diffusion model \cite{Rombach:2022:SD} through episodic training. These approaches aim to translate implicit visual understanding of VFMs into dense, pixel-level predictions.
Although this boosts in-domain results, it requires additional supervision and narrows the model scope to the training distribution (\cf \cref{fig:teaser}, \emph{orange}). 

In contrast, recent training-free approaches \cite{Liu:2023:Matcher, Zhang:2024:GF-SAM} forego task-specific training, exploiting the complementary strengths of multiple pre-trained components: DINOv2 \cite{Oquab:2023:Dinov2} for robust visual correspondence and SAM \cite{Kirillov:2023:SAM} for producing accurate masks. By relying purely on pre-trained models, these methods avoid the pitfalls of fine-tuning, achieving stronger generalization (\cref{fig:teaser}, \emph{blue}).
Nevertheless, they need to coordinate multiple VFMs, add significant computational overhead, and cannot fully exploit the intrinsic synergy between correspondence and segmentation. 

\begin{figure}[t]
        \centering
    \includegraphics[width=\linewidth]{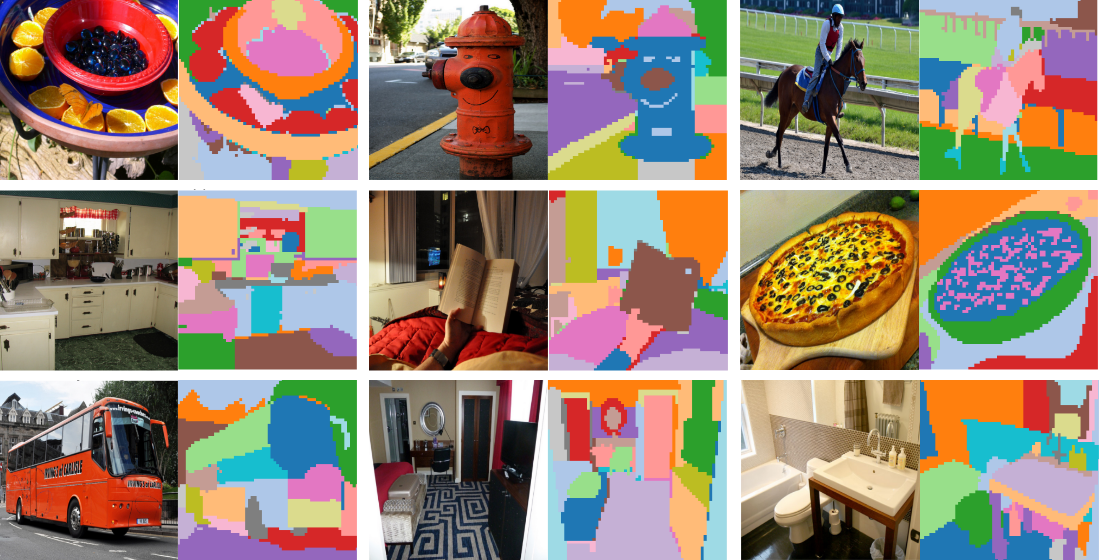}
    \vspace{-1.75em}
    \caption{\textbf{Region-level grouping from DINOv3.} Each pair shows an input image \emph{(left)} and the corresponding clustering map \emph{(right)} obtained by applying agglomerative clustering to dense DINOv3 features. The resulting clusters delineate coherent object- and part-level regions, providing a structured decomposition of the scene.}
    \label{fig:clustering}
    \vspace{-0.6em}
\end{figure}

Overall, existing ICS methods rely explicitly or implicitly on segmentation priors learned through supervision, whether from SAM pre-training or downstream fine-tuning. The recent DINOv3 model \cite{Simeoni:2025:Dinov3} may hold the key to changing this.
This purely self-supervised VFM, trained on massive-scale image corpora, is explicitly designed to produce dense localized features, unlike its predecessors \cite{Caron:2021:DINO, Oquab:2023:Dinov2}.
Its objective preserves spatial structure, enabling robust region-level grouping (\cref{fig:clustering}).
This urges us to ask if ICS can emerge directly from the DINOv3 representation, without any decoder, fine-tuning, or model composition. 

To this end, we propose \ours\ (\textbf{In}-context \textbf{S}egmentation w\textbf{I}th \textbf{D}INOv\textbf{3}), a minimalist and training-free approach, relying solely on DINOv3 features. \ours{} operates in three conceptual stages:
\emph{(i)} \emph{Fine-grained clustering} of target image features allows to obtain part-level region candidates (\cref{fig:clustering}). 
\emph{(ii)} \emph{Seed-cluster selection} identifies the most discriminative cluster through cross-image similarity between a prototype of the annotated example(s) and each cluster in the target. 
Relying on region-level similarity suppresses spurious pixel matches and resolves competition among many candidates.
\emph{(iii)} \emph{Aggregation guided by self-similarity} of DINOv3 features within the target image then merges the seed cluster with other highly affine clusters, producing a spatially coherent mask that recovers the full extent of the prompted concept. 

Finally, we uncover a subtle, yet significant limitation of correspondences from DINOv3: feature similarities across unrelated images exhibit systematic activations aligned with absolute spatial coordinates 
(\eg, features from the left side of two images tend to spuriously match regardless of semantics, as shown in \cref{fig:similarity_map}). 
This \textit{positional bias}, likely an effect of the superposition of positional encodings and semantic signals, hinders reliable correspondence reasoning in matching tasks. We propose a simple correction: we estimate the subspace affected by positional bias from a noise image and perform matching only in its orthogonal complement. This lightweight operation improves cross-image matching and, as we show, even generalizes beyond ICS.

In summary, we propose \ours{}, a principled, minimalist, yet accurate method for in-context segmentation from DINOv3 alone. It is applicable across diverse semantic granularities, \eg, from \textit{objects} to \textit{parts}, and demonstrates that emergent segmentation behavior can arise naturally from self-supervision without any training or fine-tuning. 

\smallskip\noindent Summarizing, we make the following contributions: 

\begin{itemize}
    \item We are the first to show that \emph{a self-supervised VFM suffices for training-free in-context segmentation}, building on DINOv3's core strengths of robust correspondence and its dense, localized feature structure.
    
    \item Despite its simplicity, \ours{} \emph{generalizes better across the board}, from traditional, challenging benchmarks to out-of-domain datasets and part segmentation (\cref{fig:teaser}, \emph{purple}), outperforming fine-tuned \emph{and} training-free approaches relying on SAM by an average of +\SI{7.5}{\%} mIoU.
    
    \item We unveil a \textit{positional bias in DINOv3}, which impairs its effectiveness in matching features across images, and present a simple training-free correction that generalizes beyond ICS, achieving gains of up to +\SI{6.6}{\%} PCK on the related task of semantic correspondence.
\end{itemize}

\section{Related Work}
\label{sec:related}

\begin{figure*}
        \centering
    \includegraphics[width=\linewidth]{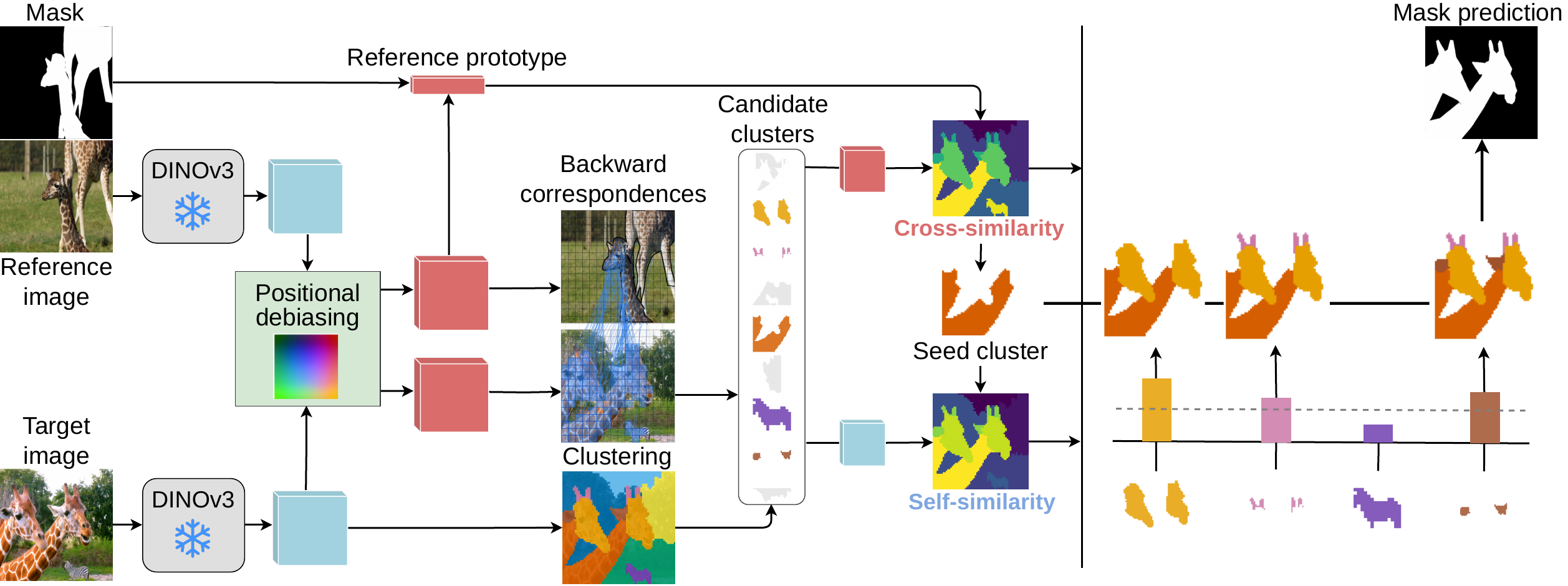}
    \caption{\textbf{Overview of \ours{}.}
We leverage the semantic and spatial structure of DINOv3 to perform in-context segmentation without training or model composition. Dense features from the reference and target images are first debiased to suppress positional bias, improving cross-image matching. The target is then decomposed into coherent regions through agglomerative clustering, providing a structured representation. 
We retain candidate clusters that match the reference through backward correspondence in the debiased space; a reference prototype derived from the annotated region anchors the \emph{seed cluster} via cross-image similarity. Finally, we combine cross-image similarity, capturing semantic alignment, with self-similarity, measuring the affinity of each cluster to the seed, to form the final mask from the seed.}
    \label{fig:method}
    \vspace{-0.2em}
\end{figure*} 

\textbf{In-context segmentation} (ICS) draws inspiration from LLMs \cite{Brown:2020:GPT, Ouyang:2022:InstructGPT,Chowdhery:2023:Palm, Touvron:2023:Llama}, which can be adapted to new tasks given contextual examples. SegGPT \cite{Wang:2023:SegGPT} and Painter \cite{Wang:2023:Painter} translate this idea to computer vision by training a generalist model to handle multiple segmentation scenarios.
Recently, this idea has been revisited in light of the advent of large-scale pre-trained VFMs: Matcher \cite{Liu:2023:Matcher} uses an annotated example to perform one-shot \emph{semantic} and \emph{part} segmentation, while PerSAM \cite{Zhang:2023:PerSAM} focuses on one-shot \emph{personalized} segmentation.
Although related in spirit to few-shot segmentation~\cite{Wang:2019:PaNet, Cuttano:2025:Sansa, Hong:2022:VAT, Lang:2022:BAM}, which learns from base classes and evaluates on disjoint novel ones defined within each dataset, ICS differs in scope and evaluation. In particular, we refer to ICS as a unified formulation of one-shot \emph{semantic}, \emph{part}, and \emph{personalized} segmentation across different levels of semantic granularity within a single, general-purpose model. 

Recent work follows two trends: \emph{Training-free pipelines} \cite{Liu:2023:Matcher, Zhang:2024:GF-SAM, Espinosa:2025:Notime} combine the semantic understanding of DINOv2 with segmentation priors from SAM, benefiting from strong generalization but inheriting SAM's mask granularity and the computational burden of multi-stage designs. \emph{Supervised methods} \cite{Meng:2024:SEGiC, Zhu:2024:Unleashing} aim to unify both capabilities within a single VFM by injecting segmentation functionality via task-specific supervision. SegIC \cite{Meng:2024:SEGiC} trains a segmentation decoder on top of DINOv2, while DiffewS \cite{Zhu:2024:Unleashing} fine-tunes Stable Diffusion \cite{Rombach:2022:SD}. Such training/fine-tuning couples the model to the training distribution, limiting its flexibility on unseen domains and granularities.
In contrast, we address ICS with a single VFM \emph{and} without training.

\myparagraphnospace{Dense self-supervised representation learning} (SSL) aims to learn dense feature extractors from unlabeled data, enabling a broad range of vision tasks~\cite{Ericsson:2021:HWD, Gui:2024:SSL}. Initial self-supervised approaches employ image-level pre-text tasks \cite{Doersch:2015:UVR, Noroozi:2016:UVR, Komodakis:2018:URL, Chen:2020:SimCLR, Grill:2020:BYOL, Caron:2020:SwAV, He:2020:MoCo, Caron:2018:DeepCluster}, transferring suboptimally to pixel-level prediction \cite{Yang:2021:InsLoc, Yang:2022:InsCon}. Later work aims to learn localized and discriminative dense features. Emergent properties in ViTs~\cite{Caron:2021:DINO} can be uncovered through spatially local objectives~\cite{Zhou:2021:iBOT, Oquab:2023:Dinov2}, 
spatio-temporal consistency~\cite{Jabri:2020:Space}, or spatial alignment across views~\cite{Pinheiro:2020:VADER, Bardes:2022:VICRegl}. 
Localized supervision is also possible through contrastive objectives on region proposals~\cite{Henaff:2021:DetCon, Henaff:2022:ODIN},  or by predicting the cluster identity of masked tokens~\cite{Darcet:2025:CPLP}.
Moreover, SSL features can be refined \emph{a-posteriori} \cite{Wysoczanska:2024:DINOiser, Hamilton:2022:USS} or through limited fine-tuning \cite{Salehi:2023:Time, Jevtic:2025:SceneDINO}.
Recent efforts distill DINOv2 \cite{Oquab:2023:Dinov2} together with weakly supervised VFMs, \eg, SAM \cite{Kirillov:2023:SAM} or CLIP \cite{radford2021clip}, to enhance spatial fidelity \cite{Heinrich:2025:RADIOv2, Ranzinger:2024:RADIO, Bolya:2025:PE}.
Most recently, DINOv3~\cite{Simeoni:2025:Dinov3} uses significant data and model scaling, a Gram anchoring objective, and high-resolution post-training to obtain an expressive, dense feature extractor. We show that dense DINOv3 features can be directly leveraged for in-context segmentation without fine-tuning or model composition.

\section{In-context Segmentation with \ours{}}
\label{sec:method}

Our goal is to segment arbitrary
concepts, \ie, objects, parts, or personalized instances, given an in-context example, using a frozen DINOv3 encoder without training or model composition.
A key property of DINOv3 is the strong self-similarity of its dense features, naturally grouping coherent parts or objects (\cref{fig:clustering}). However, in-context segmentation also requires establishing correspondences \textit{across} images, which we find affected by a systematic \textit{positional bias}: features from similar positions spuriously match across unrelated images. To address this, we propose a simple, training-free strategy to remove positional components from the features (\cref{sec:unlocking}). We use these \textit{debiased features} for cross-image matching, while retaining the original features for intra-image similarity and clustering.

Our approach, named \ours{} and illustrated in \cref{fig:method}, first partitions the target image into semantically coherent regions using self-similarity (\cref{sec:clustering}). Then it identifies the cluster that is most semantically aligned with the reference region through cross-image similarity in the debiased space (\cref{subsec:matching}). Finally, it expands this seed region by aggregating clusters according to intra-image self-similarity, yielding a complete and coherent segmentation mask (\cref{subsec:aggregation}).

\myparagraph{Task definition.}
We let $\mathbf{I}^r \in \mathbb{R}^{H \times W \times 3}$ denote the reference image with its binary mask $\mathbf{M}^r \in \{0,1\}^{H \times W}$, and $\mathbf{I}^t \in \mathbb{R}^{H \times W \times 3}$ a target image.
We extract dense features from a frozen DINOv3 encoder $\Phi(\cdot)$ \cite{Simeoni:2025:Dinov3}:
\begin{equation}
\mathbf{F}^r = \Phi(\mathbf{I}^r), \qquad \mathbf{F}^t = \Phi(\mathbf{I}^t),
\end{equation}
where $\mathbf{F}^r, \mathbf{F}^t \in \mathbb{R}^{P \times D}$ denote the $D$-dimensional patch embeddings at resolution $P = H' \times W'$.
We let $\Omega = \{1, \ldots, P\}$ denote the set of patch indices in $\mathbf{F}^r$ and $\mathbf{F}^t$.

\subsection{Unlocking the DINOv3 feature space}
\label{sec:unlocking}

\begin{figure}[t]
    \sffamily
    \centering
    \begin{minipage}{0.49\columnwidth}
        \setlength{\tabcolsep}{0.003\columnwidth}
        \begin{tabular}{cccc}
            \tiny{Reference} & \tiny{Target} & \tiny{Original} & \tiny{Debiased} \\[-1.75pt]
            \includegraphics[width=0.245\columnwidth]{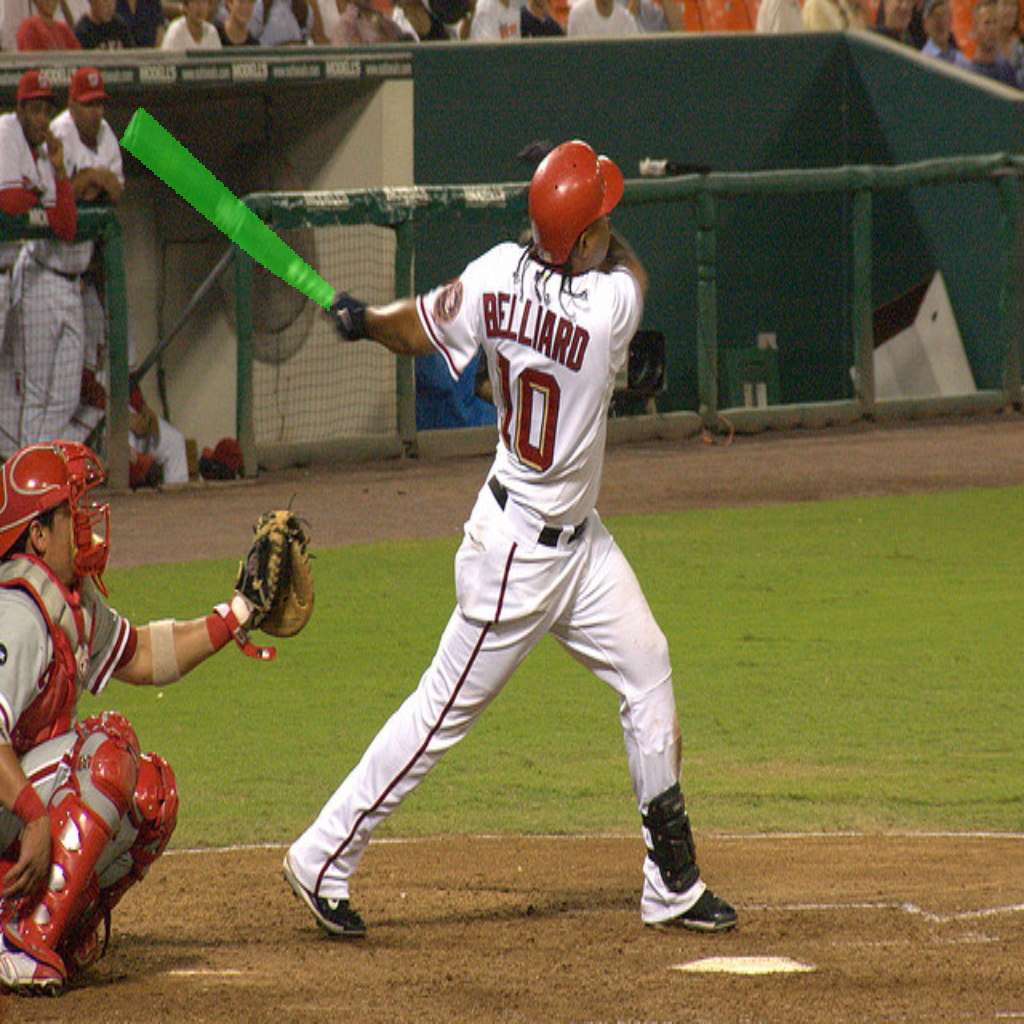} & \includegraphics[width=0.245\columnwidth]{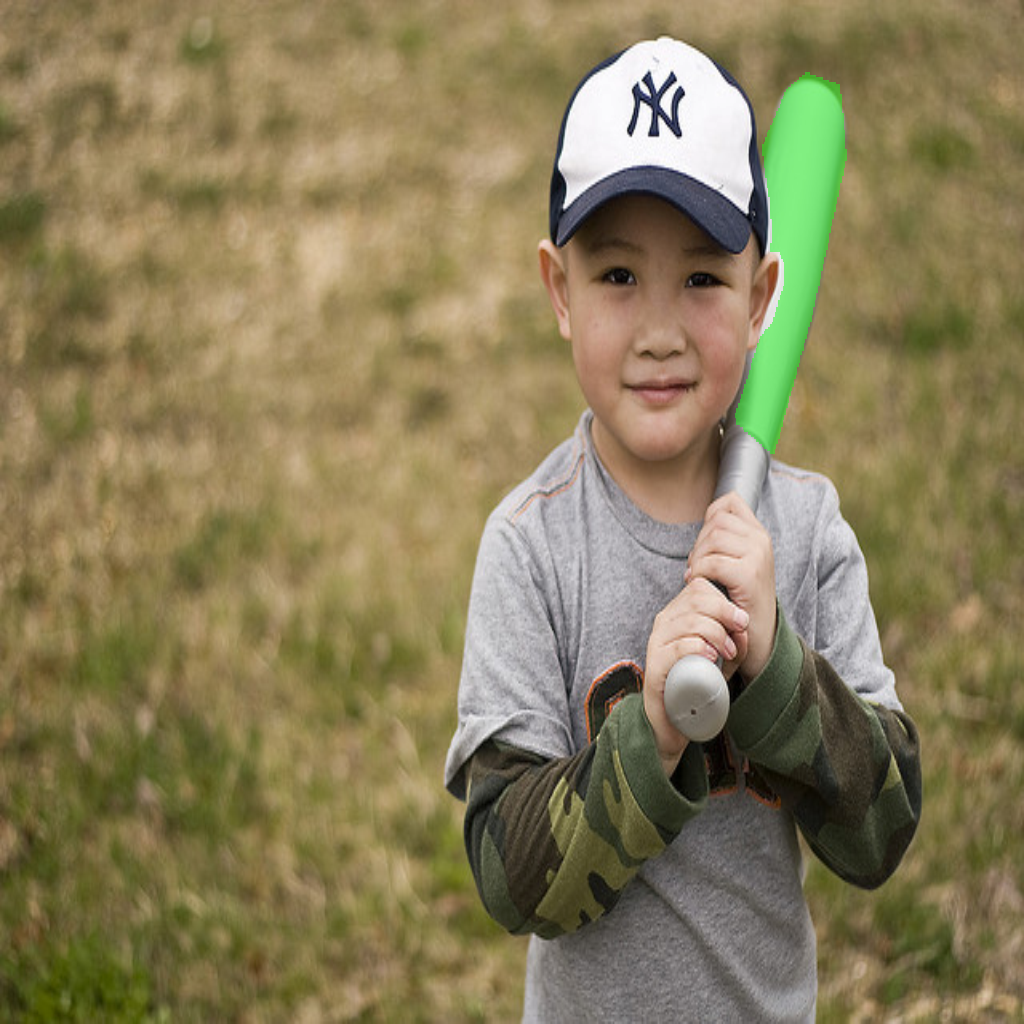} & \includegraphics[width=0.245\columnwidth]{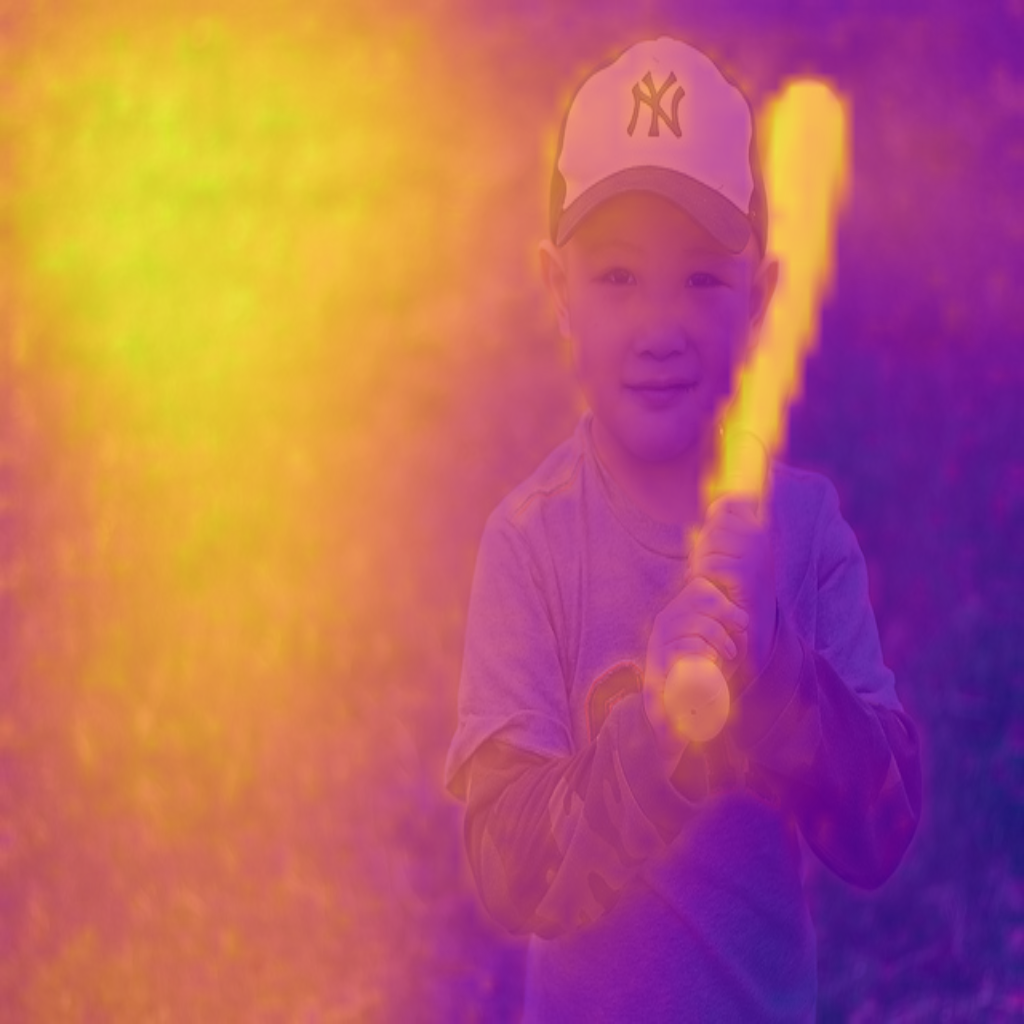} & \includegraphics[width=0.245\columnwidth]{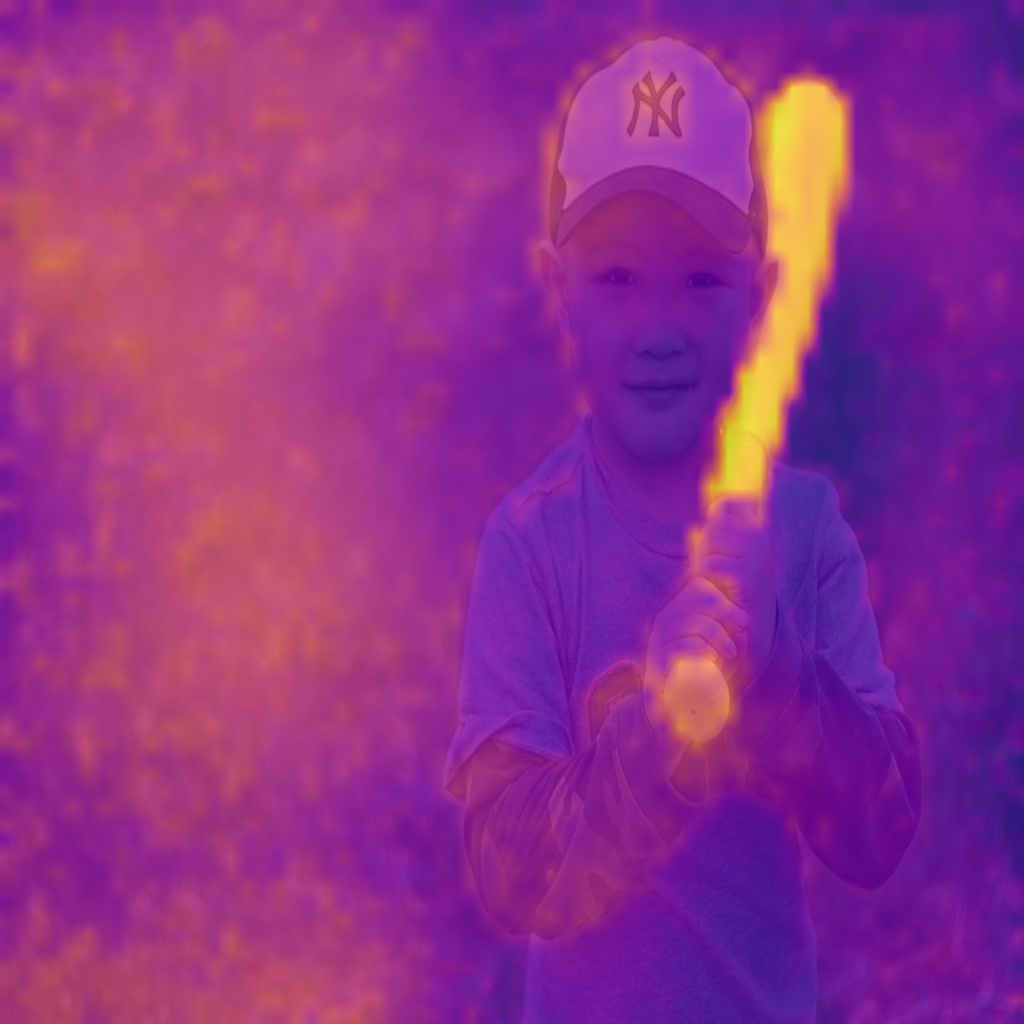} \\[-2.75pt]
            \includegraphics[width=0.245\columnwidth]{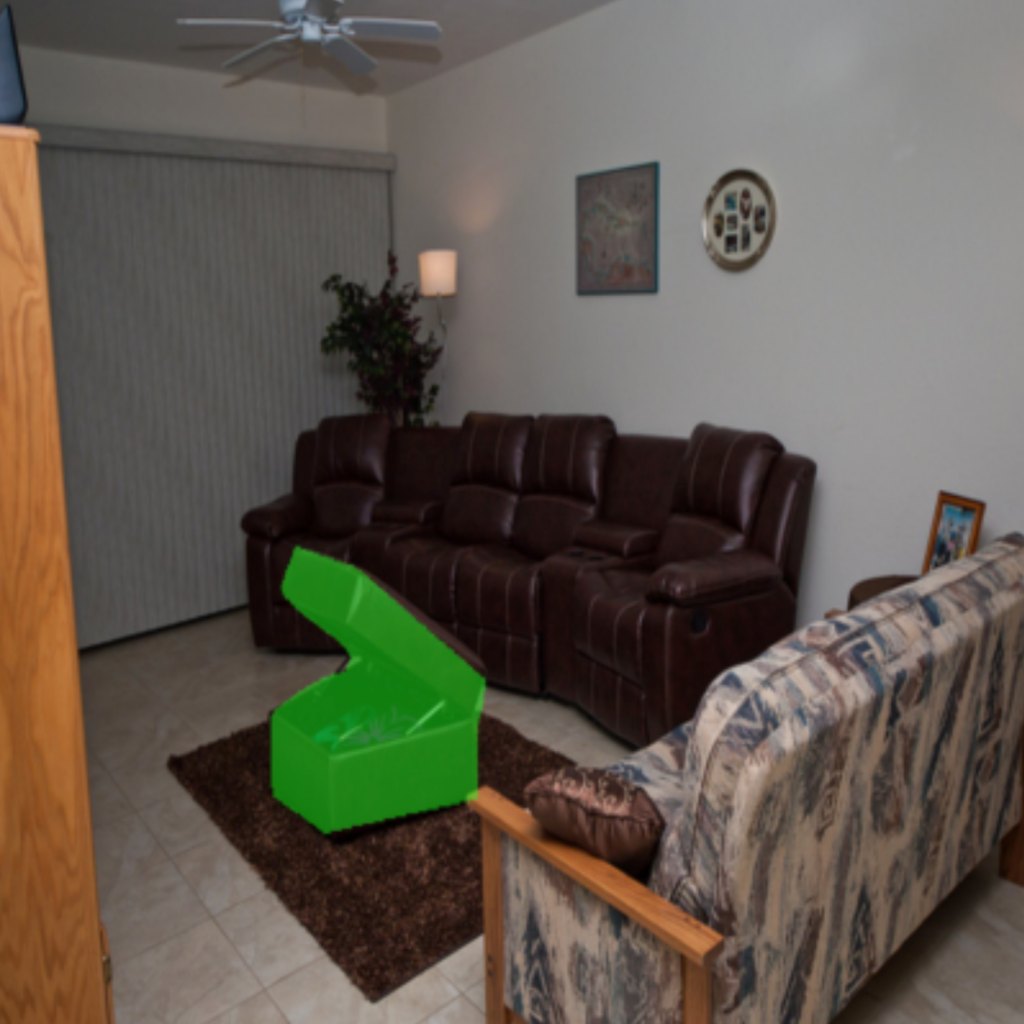} & \includegraphics[width=0.245\columnwidth]{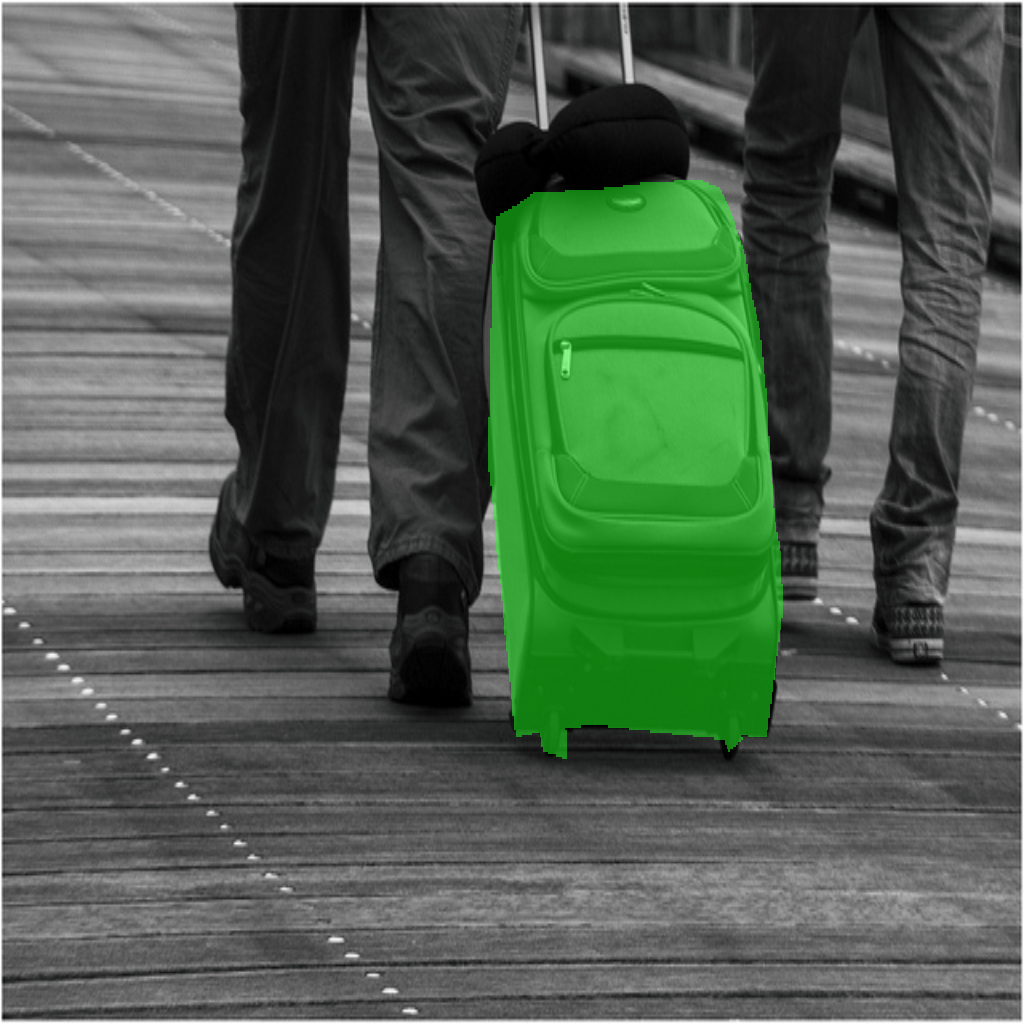} & \includegraphics[width=0.245\columnwidth]{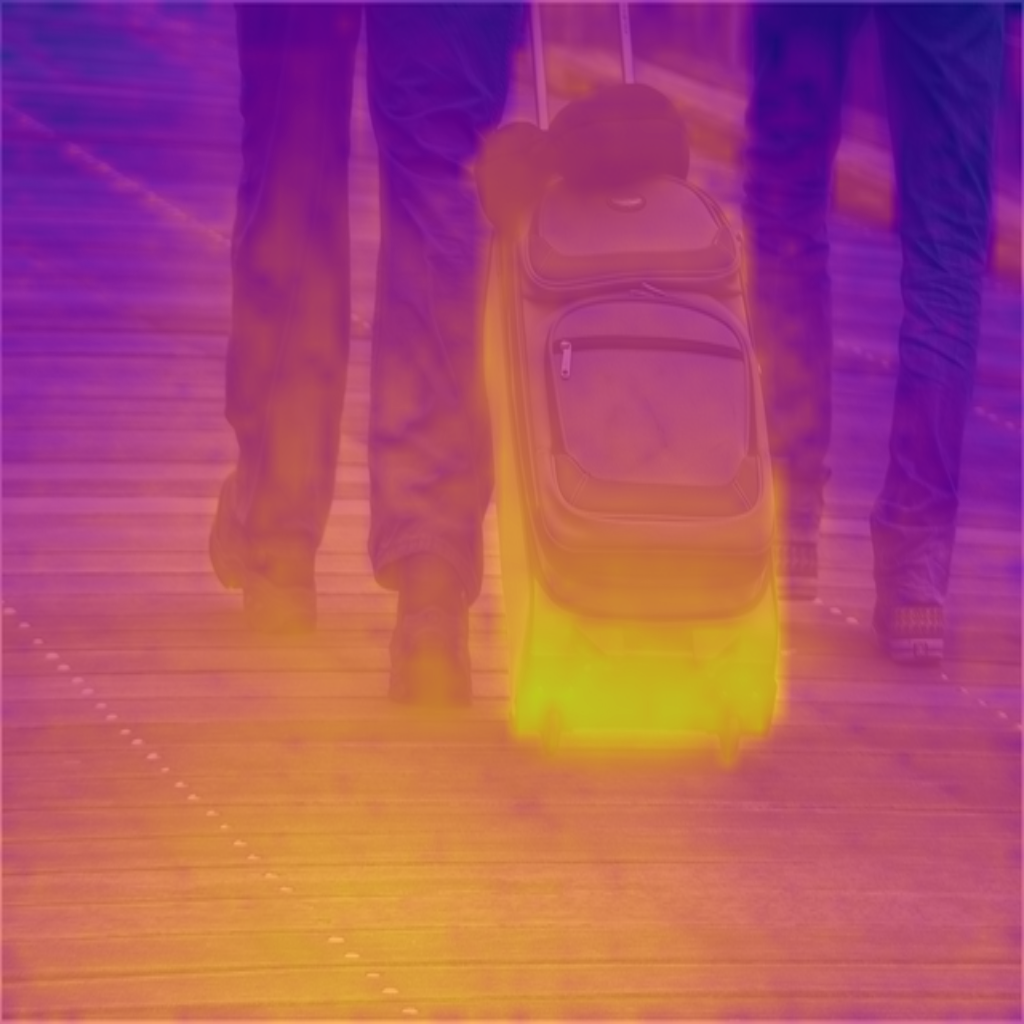} & \includegraphics[width=0.245\columnwidth]{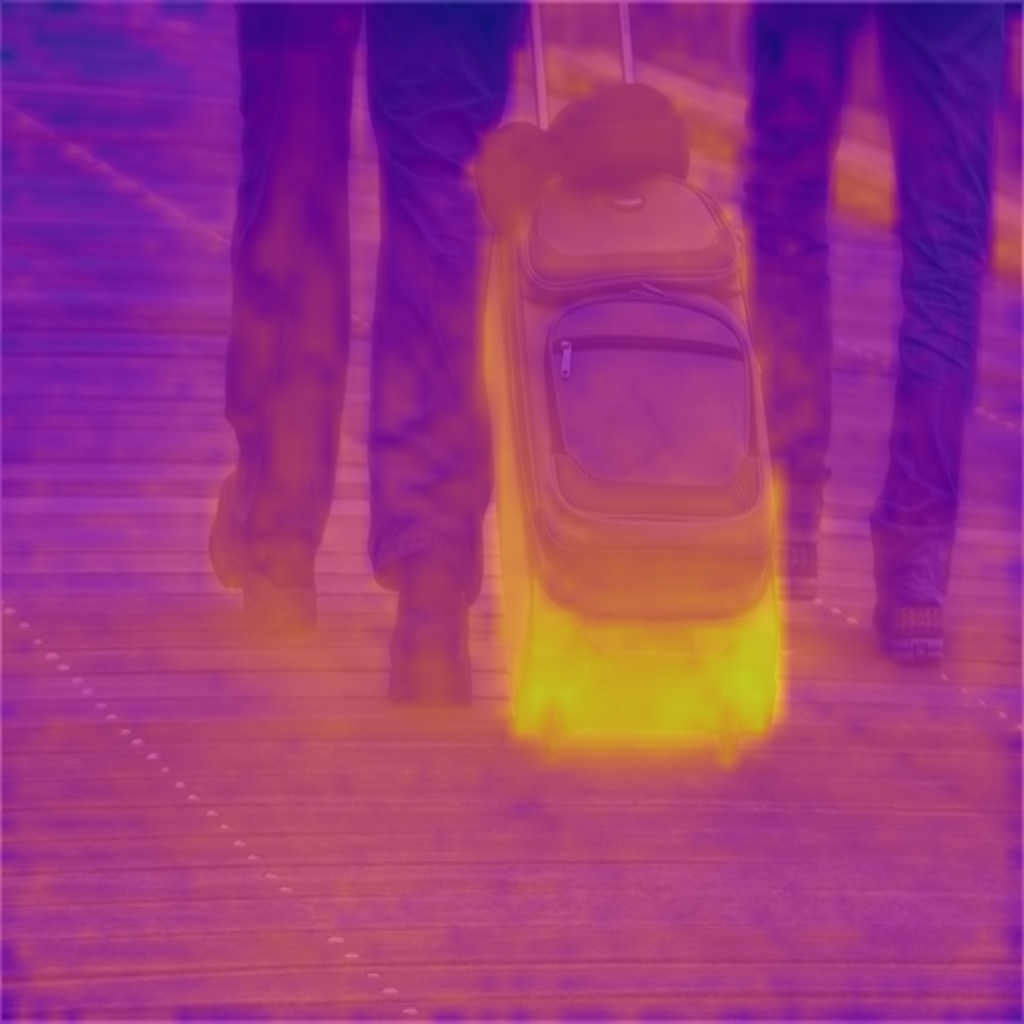}
        \end{tabular}
        \vspace{-4pt}
        \subcaption{Cross-image similarity map using an \textbf{object region} as reference.}\label{tab:objectregion}
    \end{minipage}%
    \hfill%
    \begin{minipage}{0.49\columnwidth}
        \setlength{\tabcolsep}{0.003\columnwidth}
        \begin{tabular}{cccc}
            \tiny{Reference} & \tiny{Target} & \tiny{Original} & \tiny{Debiased} \\[-1.75pt]
            \includegraphics[width=0.245\columnwidth]{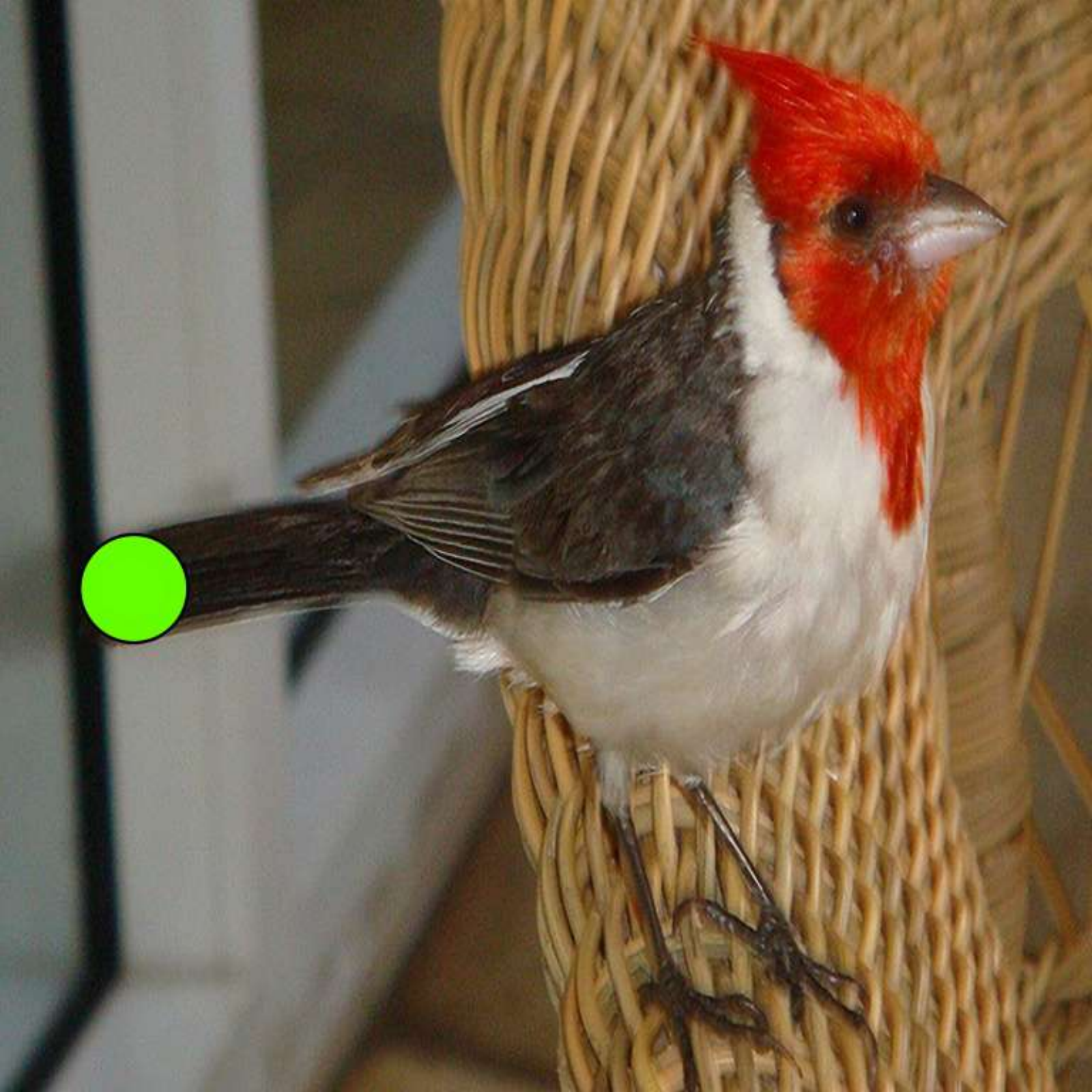} & \includegraphics[width=0.245\columnwidth]{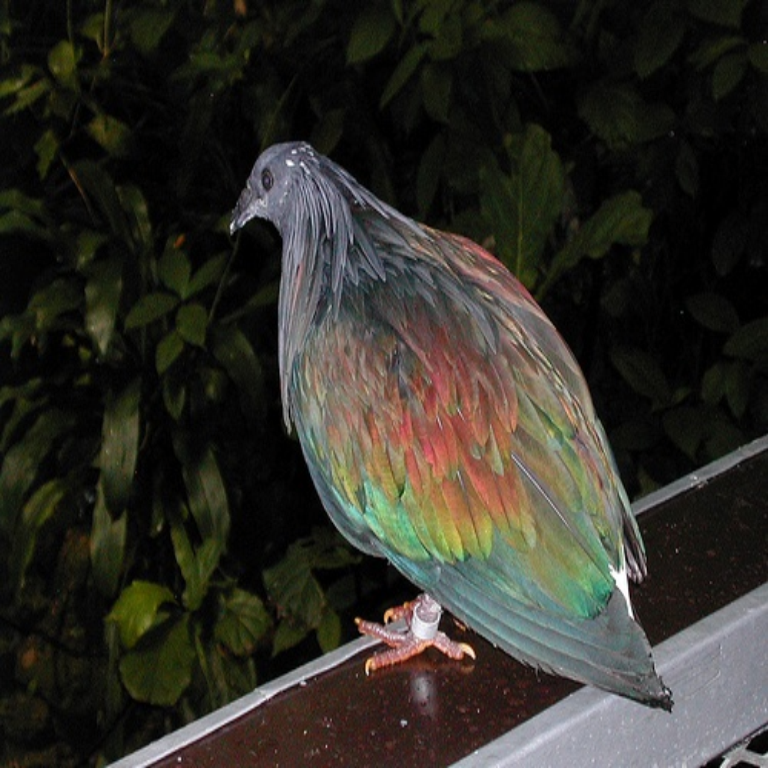} & \includegraphics[width=0.245\columnwidth]{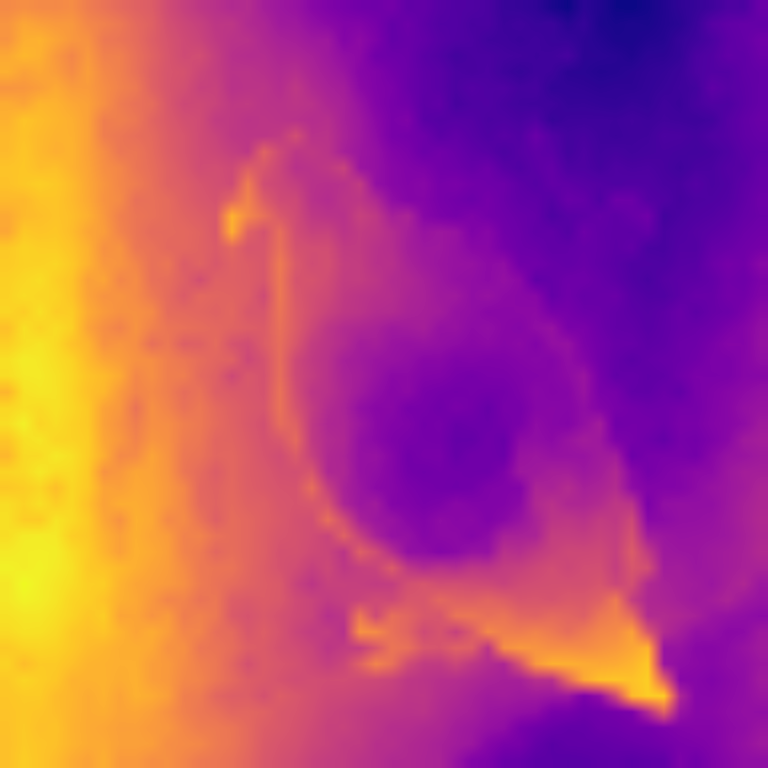} & \includegraphics[width=0.245\columnwidth]{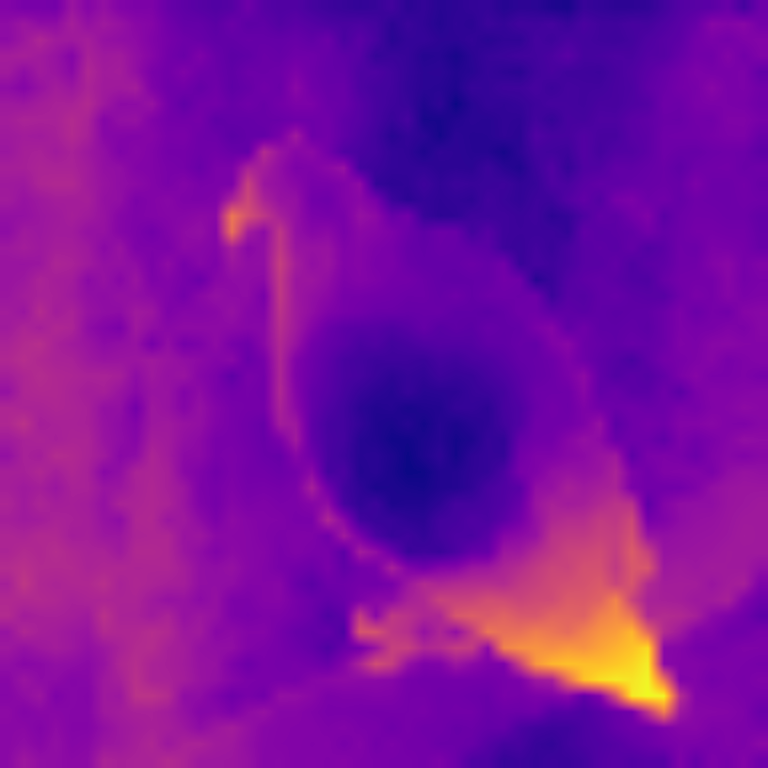} \\[-2.75pt]
            \includegraphics[width=0.245\columnwidth]{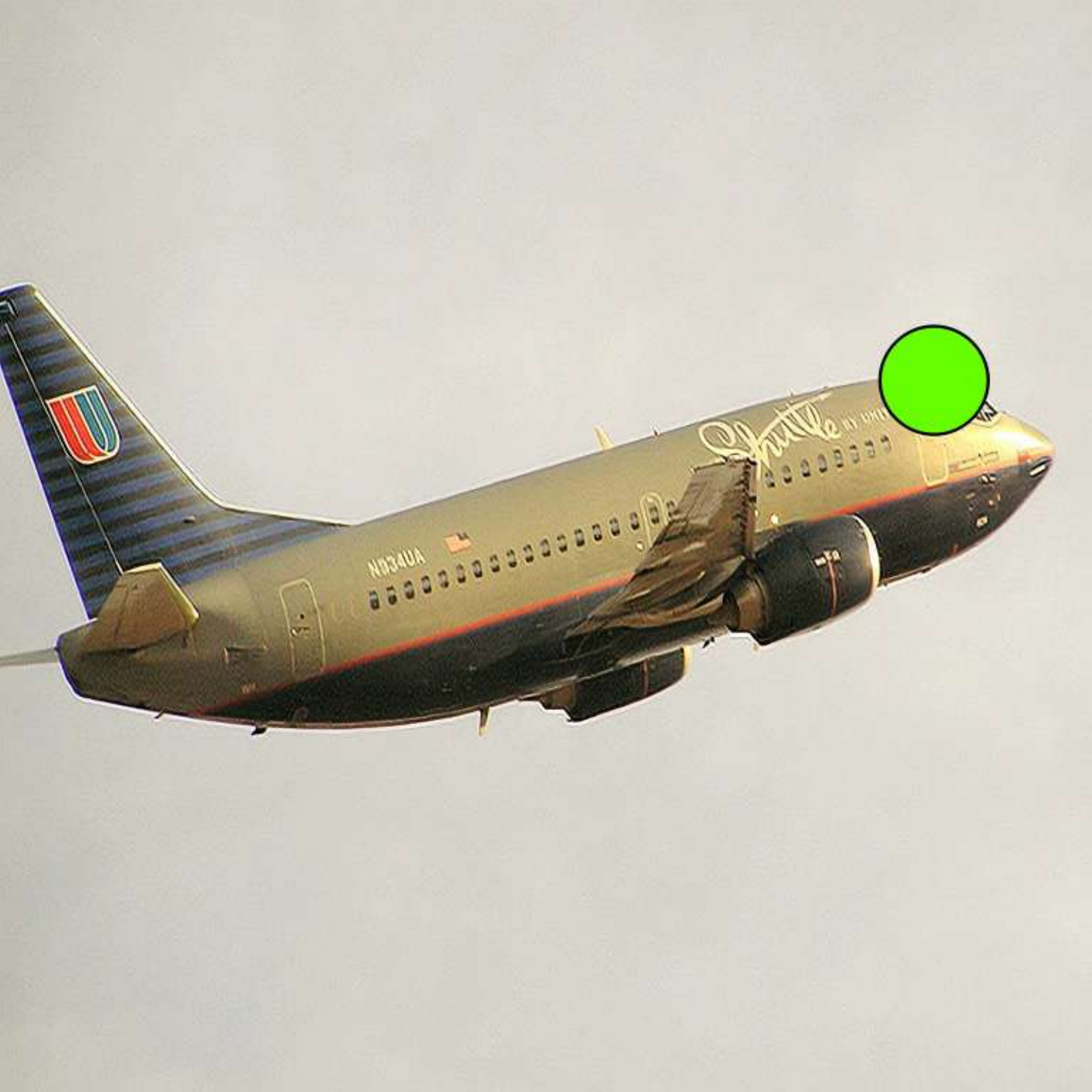} & \includegraphics[width=0.245\columnwidth]{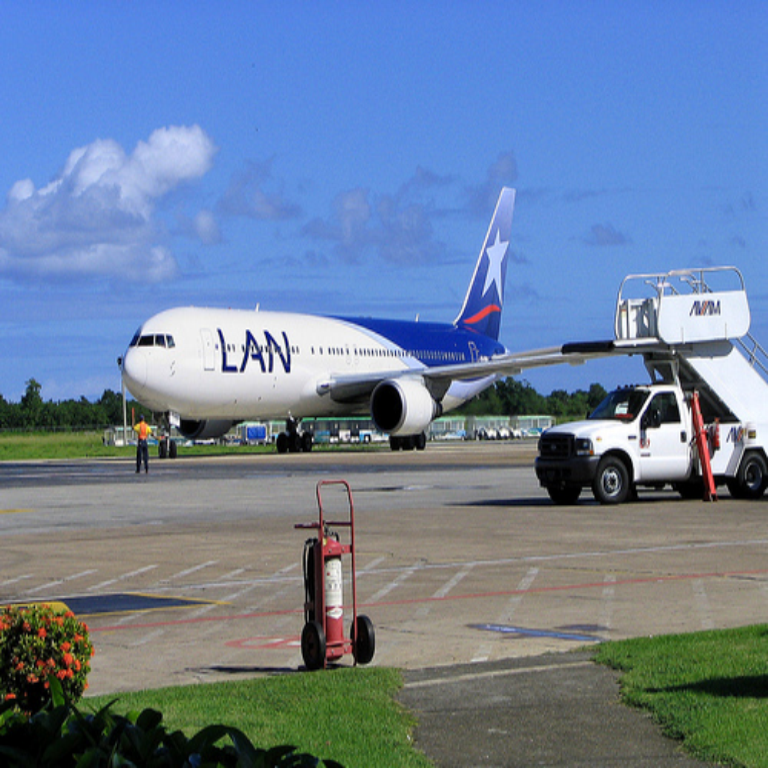} & \includegraphics[width=0.245\columnwidth]{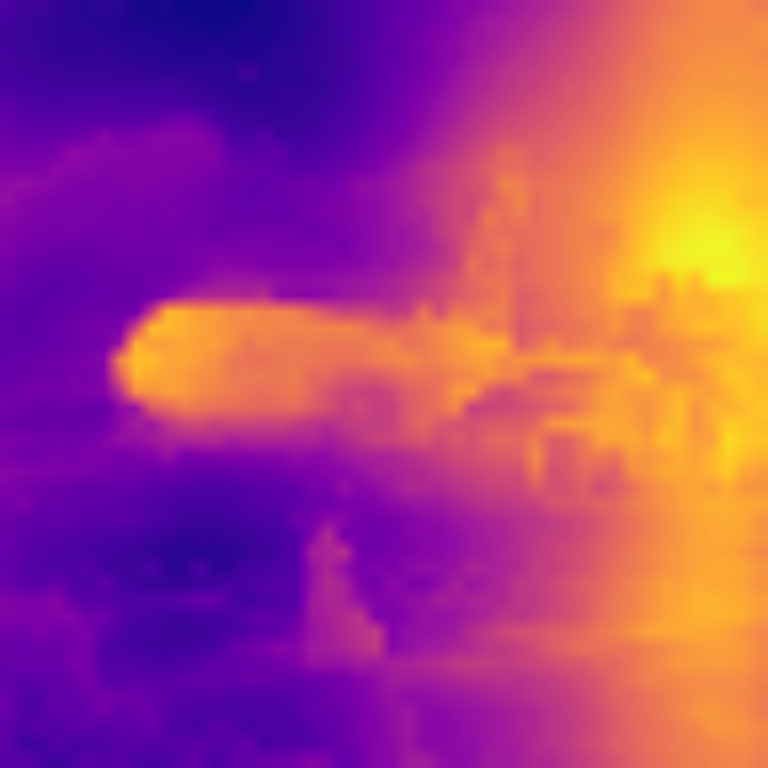} & \includegraphics[width=0.245\columnwidth]{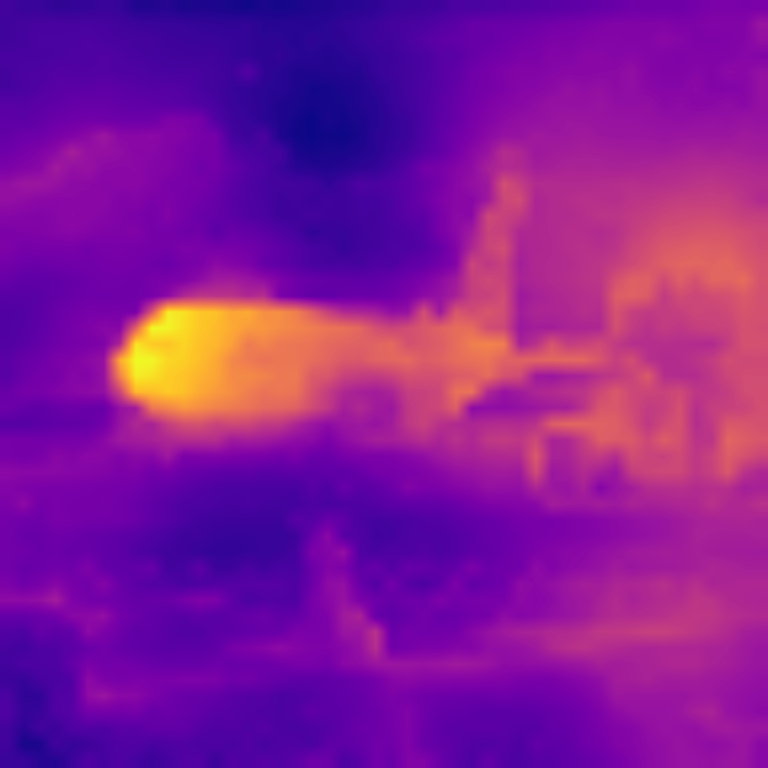}
        \end{tabular}
        \vspace{-4pt}
        \subcaption{Cross-image similarity map using a \textbf{keypoint} as reference.}\label{tab:keypoint}
    \end{minipage}\\
    \vspace{3pt}
    \begin{minipage}{1.0\columnwidth}
        \setlength{\tabcolsep}{0.00335\columnwidth}
        \hspace{-0.35pt}\begin{tabular}{ccccccccc}
            \includegraphics[width=0.105\columnwidth]{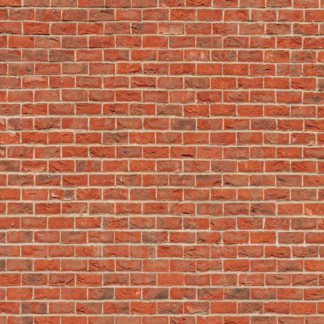} & \includegraphics[width=0.105\columnwidth]{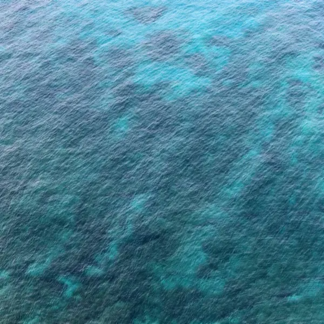} & \includegraphics[width=0.105\columnwidth]{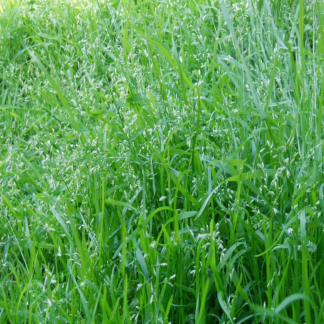} & \includegraphics[width=0.105\columnwidth]{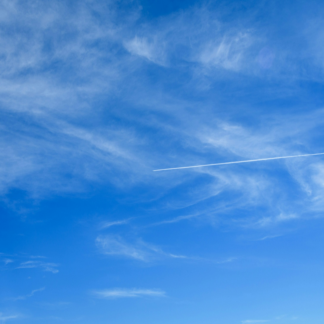} & \includegraphics[width=0.105\columnwidth]{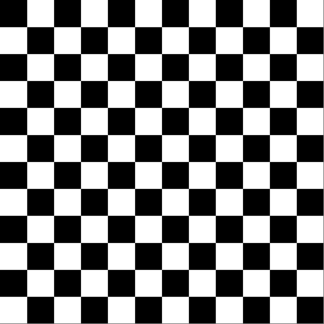} & \includegraphics[width=0.105\columnwidth]{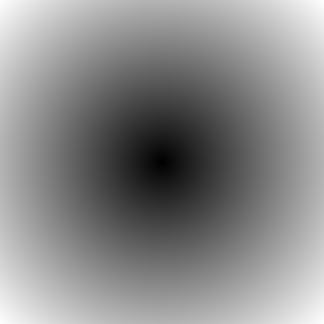} & \includegraphics[width=0.105\columnwidth]{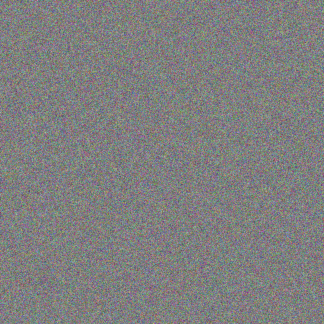} & \includegraphics[width=0.105\columnwidth]{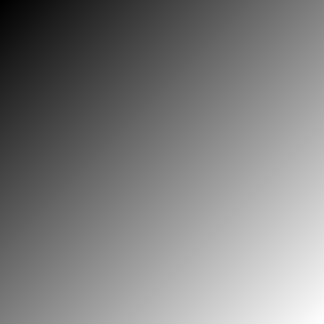} & \includegraphics[width=0.105\columnwidth]{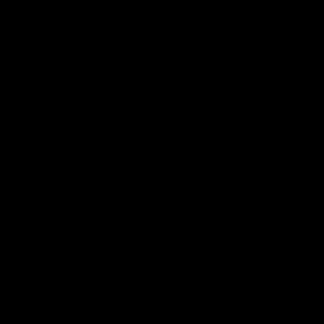} \\[-2.1pt]
            
            \includegraphics[width=0.105\columnwidth]{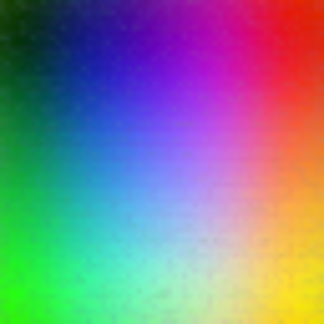} & \includegraphics[width=0.105\columnwidth]{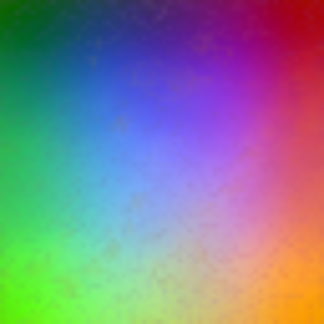} & \includegraphics[width=0.105\columnwidth]{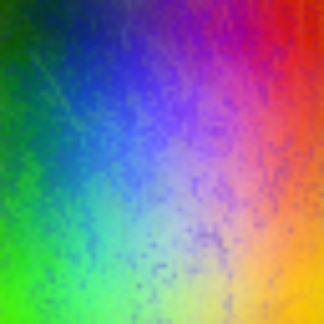} & \includegraphics[width=0.105\columnwidth]{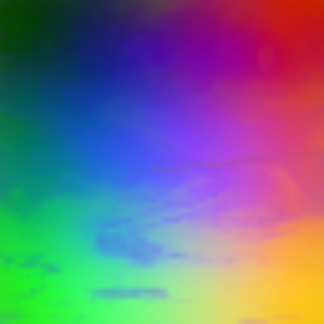} & \includegraphics[width=0.105\columnwidth]{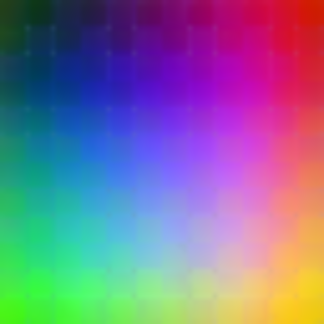} & \includegraphics[width=0.105\columnwidth]{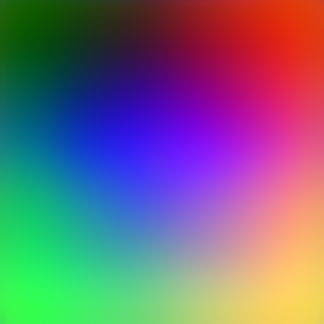} & \includegraphics[width=0.105\columnwidth]{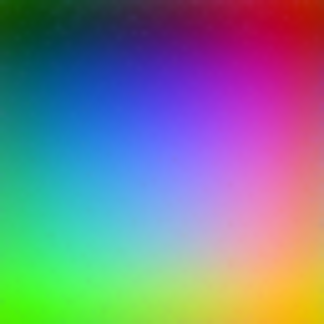} & \includegraphics[width=0.105\columnwidth]{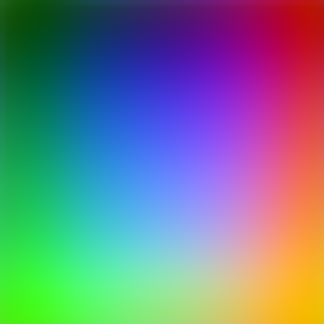} & \includegraphics[width=0.105\columnwidth]{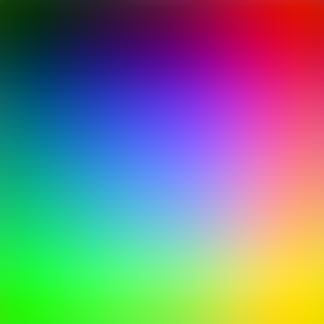} \\
        \end{tabular}
        \vspace{-4pt}
        \subcaption{\textbf{Positional subspace.} PCA on uniform and low-complexity images.}\label{tab:pospca}
    \end{minipage}
    \vspace{-0.3em}
    \caption{\textbf{Positional bias in DINOv3 features.} For both region \emph{\subref{tab:objectregion}} and keypoint \emph{\subref{tab:keypoint}} prompts, similarity maps computed with the original DINOv3 features show structured activations aligned with the reference coordinates, independent of semantics. Our debiased features mitigate this behavior. \emph{\subref{tab:pospca}} PCA of features from images with low semantic complexity (\eg, noise, flat textures) reveals a stable low-dimensional positional subspace underlying this bias.\label{fig:similarity_map}}
\vspace{-0.5em}
\end{figure}

Solving the ICS task fundamentally relies on computing robust and reliable feature correspondences between the reference and target images \cite{Liu:2023:Matcher}.
As a diagnostic tool to evaluate DINOv3's ability to establish reliable correspondences, we compute cross-image similarity to visualize how target patches align with the reference concept.
Specifically, given the reference mask $\mathbf{M}^r$ and the set of foreground patch indices\footnote{In slight abuse of notation, we let $\mathbf{M}^r_j$ denote the $j^\text{th}$ patch of $\mathbf{M}^r$.} $\mathcal{R} = \{ j \in \Omega \mid \mathbf{M}^r_j = 1 \}$,
we compute a reference prototype $\mathbf{p}^r$ and its similarity to each target patch $i\in\Omega$:
\begin{equation}
    \label{eq:naive_prototype}
    \mathbf{p}^r = \frac{1}{|\mathcal{R}|} \sum\nolimits_{j \in \mathcal{R}} \mathbf{F}^r_j, \qquad \text{sim}(i) = \langle \mathbf{F}^t_i, \mathbf{p}^r \rangle.
\end{equation}
This produces dense similarity maps, indicating how well each target patch aligns with the reference concept. We visualize these maps at two granularity levels: \textit{(i)} at mask level (\cref{fig:similarity_map}a), where the reference corresponds to an object, and \textit{(ii)} at keypoint level (\cref{fig:similarity_map}b), where the reference is a single annotated keypoint.
The resulting similarity maps show that DINOv3 captures meaningful semantic correspondences between reference and target. 
However, they also exhibit a stable \textit{positional bias}: features at a given position in the reference tend to produce spurious activations at the same position in the target, irrespective of semantics.
These false activations typically occur where the target area lacks semantic content (\eg, uniform background regions), suggesting that positional information dominates weak semantic cues.
To ground this intuition, \cref{fig:similarity_map}c visualizes features from inputs with minimal semantic content: a principal component analysis (PCA) suggests a stable low-dimensional subspace associated with positional signals.  

We use this signal as a simple and effective approximation of positional bias, which can be estimated once and removed consistently at inference time.
Specifically, we estimate the positional subspace by passing a noise image $\mathbf{I}^{\text{noise}} \sim \mathcal{N}(\mathbf{0},\mathbf{1}) \in  \mathbb{R}^{H \times W \times 3}$ through the encoder:
\begin{equation}
\label{eq:noise_feats}
\mathbf{F}^{\text{noise}} = \Phi(\mathbf{I}^{\text{noise}}) \in \mathbb{R}^{P \times D}.
\end{equation}
We apply singular value decomposition $\mathbf{F}^{\text{noise}} = \mathbf{U}\boldsymbol{\Sigma}\mathbf{V}^\top$ and select the top $s$ right singular vectors $\mathbf{B} = \mathbf{V}_{[:, 1:s]}$ as a basis for the positional subspace.
We then project both reference and target features onto its orthogonal complement:
\begin{equation}
\tilde{\mathbf{F}}^r = \mathbf{F}^r \bigl( \mathbf{1}_D - \mathbf{B}\mathbf{B}^\top \bigr) \qquad 
\tilde{\mathbf{F}}^t = \mathbf{F}^t \bigl( \mathbf{1}_D - \mathbf{B}\mathbf{B}^\top \bigr).
\label{eq:debiasing}
\end{equation}

The effect of this projection is to suppress positional components: using these debiased features to recompute the similarity map as in \cref{eq:naive_prototype} yields activations that are less affected by structured positional bias (\textit{cf.} \cref{fig:similarity_map}).
In the rest of the paper, we refer to $\tilde{\mathbf{F}}^r,\tilde{\mathbf{F}}^t$ as \textit{debiased features}.

Interestingly, this spatial dependency is markedly weaker in DINOv2; positional correlations are less pronounced and not as easily observable in similarity maps (\cf Supp.\ Material).
We hypothesize this positional bias to be a by-product of the stronger local-consistency constraints in DINOv3. Namely, the \textit{Gram anchoring} constrains the covariance matrix of patch embeddings, encouraging global statistics of features to remain stable throughout training. While improving spatial consistency, this objective may inadvertently amplify absolute spatial correlations, resulting in residual positional bias when semantic content is weak.

\myparagraph{Unlocked feature space.}
We exploit the complementary nature of DINOv3 features by using:
\textit{(i)} our \emph{debiased features} for cross-image semantic matching, where positional signals are harmful, and
\textit{(ii)} \emph{original features} for intra-image grouping, where spatial structure is helpful.

\subsection{Fine-grained clustering}
\label{sec:clustering}

The first step of our ICS pipeline is to partition the target image into semantically coherent regions.
The dense feature maps of DINOv3 exhibit strong local consistency: As shown by \cite{Simeoni:2025:Dinov3}, patches belonging to the same object or part tend to have highly similar embeddings.
We leverage this to group image regions in an unsupervised manner.

While $K$-means clustering has been widely used in self-supervised representation learning \cite{Caron:2020:SwAV, Melas:2022:Spectral, Hamilton:2022:USS, Oquab:2023:Dinov2}, it requires predefining the number of clusters, which is ill-suited for the open-world nature and variable granularity of ICS.
Density-based methods, \eg, DBSCAN \cite{Ester:1996:DBSCAN}, struggle in high-dimensional feature spaces where the notion of density becomes unreliable \cite{Schubert:2017:DBSCAN, Radovanovic:2010:Density}, and typically require dimensionality reduction.
Instead, we adopt agglomerative clustering \cite{Mullner:2011:Agglomerative}, which progressively merges locally similar features in a bottom-up manner, naturally aligning with the spatial smoothness of DINOv3.
A single threshold hyperparameter $\tau$ provides intuitive control over the resulting granularity without fixing a predefined number of regions.

Concretely, we partition the (original) target patch embeddings $\mathbf{F}^t$ into $K$ disjoint spatial regions via iterative agglomerative clustering, yielding clusters $\{\mathcal{G}_1, \ldots, \mathcal{G}_K\}$ such that
\begin{equation}
\bigcup_{k=1}^K \mathcal{G}_k =\Omega, \qquad \mathcal{G}_i \cap \mathcal{G}_j = \emptyset \quad \forall i \neq j.
\end{equation}
As shown in \cref{fig:clustering}, this unsupervised approach produces spatially coherent clusters that provide a strong structural representation of the image.

\subsection{Seed-cluster selection}
\label{subsec:matching}

Having partitioned the target image into semantically coherent regions, we identify the cluster that best corresponds to the reference region.
We do this in two stages: \emph{candidate localization} and \emph{seed-cluster selection}.

\myparagraph{Candidate localization.}  
Directly correlating the reference prototype with all target patches, as in \cref{tab:objectregion}, often produces broad activations on related concepts, even when the reference depicts only a specific part: \eg, the prototype of a \textit{person head} may trigger responses over the full person.

To adapt matching at the correct level of granularity, we instead compute \textit{backward} correspondences, \ie for each target patch $i$, we find its most similar reference patch $j$:
\begin{equation}
\text{NN}(i) = \argmax_{j \in \Omega}
\langle \tilde{\mathbf{F}}^t_i, \tilde{\mathbf{F}}^r_j \rangle,
\label{eq:nn}
\end{equation}
Backward matching of target patches allows us to \emph{implicitly leverage unannotated negatives} in the reference image.
By retaining only target patches whose nearest neighbor in the reference falls within the support mask, we obtain a filtering mechanism that conservatively estimates the set $\mathcal{C}_{\text{NN}}$ of target patches in which the reference concept may appear as
\begin{align}
\mathcal{C}_{\text{NN}} =&
\bigl\{ i \,\big|\, \mathbf{M}^r_{\text{NN}(i)} = 1 \bigr\}.
\end{align}
Restricting the precomputed clusters $\{\mathcal{G}_k\}$ to those that overlap with $\mathcal{C}_{\text{NN}}$ yields the subset of candidate clusters
\begin{align}
\mathcal{C}_{\text{cand}} =&
\bigl\{ \mathcal{G}_k \,\big|\, \mathcal{G}_k \cap \mathcal{C}_{\text{NN}} \neq \emptyset \bigr\}.
\end{align}

\begin{table*}[t]

\caption{\textbf{Comparison of \ours{} (mIoU in \%, $\uparrow$) on one-shot semantic, part, and personalized segmentation.} 
State-of-the-art methods are grouped into {task-specific fine-tuning} and {training-free} approaches. 
Previous training-free methods rely on SAM, pre-trained with mask-level supervision, whereas \ours{} uses only frozen self-supervised DINOv3 features. 
\textcolor{indomain}{Gray} indicates the model was trained on the corresponding train split of the dataset; best results \textbf{bold}, \nth{2} best \underline{underlined}.
$\dagger$ denotes a GF-SAM variant using DINOv3 features.
}

\label{tab:main_comparison}
\vspace{-5pt}
\small
\setlength\tabcolsep{2.8pt}
\begin{tabularx}{\linewidth}{@{}Xlc|cccccc|cc|cc@{}}
\toprule
& &  & \multicolumn{6}{c}{\textbf{Semantic}} & \multicolumn{2}{c}{\textbf{Part}} & \spillC{.3cm}{\textbf{Personalized}} & \\
\textbf{Method} & \textbf{Encoder} & \textbf{\#{}Param} & 
 \textbf{LVIS-92\textsuperscript{i}}  &
 \textbf{COCO-20\textsuperscript{i}} &
 \textbf{ISIC} &
 \textbf{SUIM} & 
 \textbf{iSAID} &
 \textbf{X-Ray} &
 \textbf{PASCAL} &
 \textbf{PACO} &
 \multicolumn{1}{c|}{\textbf{PerMIS}}  & \textbf{Avg}\\
\midrule

\multicolumn{5}{@{}l}{\textbf{Task-specific fine-tuning}: \textit{Semantic + mask supervision}} \\
~~Painter \cite{Wang:2023:Painter} & \small{ViT} & \SI{354}{M} & 10.5 & \ind{33.1} & -- & -- & -- & -- & 30.4 & 14.1 & \multicolumn{1}{c|}{--} & -- \\ 
~~SegGPT  \cite{Wang:2023:SegGPT} & \small{ViT} & \SI{354}{M} & 18.6 & \ind{56.1} & 37.5 & 34.9 & \ind{30.9} & \textbf{87.5} & 35.8 & \ind{13.5} & \multicolumn{1}{c|}{18.7} & 37.1 \\ 
~~SINE   \cite{Liu:2024:SINE} & \small{DINOv2} & \SI{373}{M} & 31.2 & \ind{64.5} & 25.8 & 50.7 & 38.3 & 39.8 & 36.2 & 23.3 & \multicolumn{1}{c|}{42.5} &  39.1 \\ 
~~DiffewS \cite{Zhu:2024:Unleashing} & \small{Stable Diffusion} & \SI{890}{M} & 31.4 & \ind{71.3} & 27.8 & 48.9 & 47.5 & 41.6 & 34.0 & 22.8 & \multicolumn{1}{c|}{35.2} & 40.1 \\ 
~~SegIC \cite{Meng:2024:SEGiC}  & \small{DINOv2} & \SI{310}{M} & \ind{44.6} & \ind{76.1} & 25.3 & 52.5 & 46.1 & 34.5 & 39.9 & 25.9 & \multicolumn{1}{c|}{51.8} &  44.1 \\ 
~~SegIC\,{\scriptsize (COCO)}\,\cite{Meng:2024:SEGiC} & \small{DINOv2} & \SI{310}{M} & \underline{35.7} & \ind{75.6} & 22.5 & 52.9 & 40.8 & 30.8 & 38.6 & 25.1 & \multicolumn{1}{c|}{44.9} & 40.8 \\ 
\midrule

\multicolumn{4}{@{}l}{\textbf{Training free}: \textit{Mask-supervised pre-training}} \\
~~PerSAM \cite{Zhang:2023:PerSAM} & \small{SAM} & 640 M & 11.5 & 23.0 & 23.9 & 28.7 & 19.2 & 31.7 & 32.5 & 22.5 & \multicolumn{1}{c|}{48.6} & 26.8  \\ 
~~Matcher \cite{Liu:2023:Matcher} & \small{DINOv2 + SAM} & \SI{945}{M} & 33.0 & 52.7 & 38.6 & 44.1 & 33.3 & 70.8 & 42.9 & 34.7 & \multicolumn{1}{c|}{\underline{63.8}} & 46.0  \\ 
~~GF-SAM \cite{Zhang:2024:GF-SAM} & \small{DINOv2 + SAM} & \SI{945}{M} & 35.2 & \textbf{58.7} & 48.7 & \underline{53.1} & 47.1 & 51.0 & 44.5 & \underline{36.3} & \multicolumn{1}{c|}{54.1} & 47.6  \\ 
~~GF-SAM\textsuperscript{$\dagger$} \cite{Zhang:2024:GF-SAM} & \small{DINOv3 + SAM} & \SI{945}{M} & 31.8 & 54.8 & 50.9 & 50.5 & 46.7 & 56.1 & 44.9 & 34.4 & \multicolumn{1}{c|}{52.6} & 47.0  \\ 
~~ \tikz[baseline=1ex]{\draw[->, thick] (0., 2.8ex) -- (0,1.4ex) -- (1em,1.4ex);} + \textit{our debias} & \small{DINOv3 + SAM} & \SI{945}{M} & 34.6 & 55.9 & \underline{51.8} & 52.9 & \underline{47.6} & 60.0 & \underline{46.2} & 36.1 & \multicolumn{1}{c|}{54.5} & \underline{48.8}  \\ 
\midrule

\multicolumn{13}{@{}l}{\textbf{Training free}: \textit{Unsupervised pre-training}} \\
~~\textbf{\ours} \emph{(ours)}   & {DINOv3}       & \textbf{\SI{304}{M}} & \textbf{41.8} & \underline{57.6} & \textbf{54.4} & \textbf{54.9} & \textbf{52.1} & \underline{78.8} &  \textbf{50.5} & \textbf{38.7} & \multicolumn{1}{c|}{\textbf{67.0}} & \textbf{55.1} \\
\bottomrule
\end{tabularx}
\vspace{-0.1em}
\end{table*}

\myparagraph{Seed selection.}  
We compute prototypes in the debiased feature space for both the candidate clusters $\mathcal{G}_k\in\mathcal{C}_\text{cand}$ and the annotated reference region:
\begin{equation}
\tilde{\mathbf{p}}^t_k = \frac{1}{|\mathcal{G}_k|} 
\sum_{i \in \mathcal{G}_k} \tilde{\mathbf{F}}^t_i
\qquad
\tilde{\mathbf{p}}^r = \frac{1}{|\mathcal{R}|} 
\sum_{j \in \mathcal{R}} \tilde{\mathbf{F}}^r_j.
\end{equation}
We then compute a cross-image similarity score
\begin{equation}
s^{\text{cross}}_k = \langle \tilde{\mathbf{p}}^t_k, \tilde{\mathbf{p}}^r \rangle,
\label{eq:cross-similarity}
\end{equation}
measuring how well each candidate $\mathcal{G}_k$ aligns semantically with the reference.
The final seed cluster is selected as
\begin{equation}
\mathcal{G}^* = 
\argmax_{\mathcal{G}_k \in \mathcal{C}_{\text{cand}}} s^{\text{cross}}_k,
\end{equation}
corresponding to the target region that is most semantically aligned with the reference at the correct part granularity.

\subsection{Cluster aggregation}
\label{subsec:aggregation}

The seed cluster $\mathcal{G}^*$ provides a strong but typically \emph{partial} localization of the semantic concept in the target, often covering only the most discriminative part of the concept, such as a person's head or the neck of a giraffe (\cf \cref{fig:method}).
To recover the full extent of the concept, we evaluate all remaining candidate clusters to decide which should be merged.
Intuitively, the cross-image similarity score $s^{\text{cross}}_k$ (Eq.~\ref{eq:cross-similarity}), reflects how semantically close candidate clusters $\mathcal{G}_k$ are to the reference. 
However, relying solely on cross-image similarity can be unreliable under occlusions or viewpoint changes, where semantically relevant regions may appear dissimilar.
Therefore, we propose to complement \emph{semantic alignment} (across images) with \emph{structural coherence} (within the target image). Specifically, we exploit a key property of {DINOv3}~\cite{Simeoni:2025:Dinov3}: \emph{its features exhibit strong self-similarity within the same image}.  
Hence, clusters belonging to the same concept tend to lie close in feature space.
For each candidate cluster, we thus compute its similarity to the seed in the \emph{original} feature space as
\begin{equation}
\bar{\mathbf{p}}^t_k = \frac{1}{|\mathcal{G}_k|} 
\sum_{i \in \mathcal{G}_k} \mathbf{F}^t_i, 
\qquad
s^{\text{intra}}_k = \langle \bar{\mathbf{p}}^t_k, \bar{\mathbf{p}}^t_* \rangle,
\label{eq: seed_cluster_prot}
\end{equation}
where $\bar{\mathbf{p}}^t_*$ denotes the prototype of the seed cluster $\mathcal{G}^*$.

\myparagraph{Final aggregation.}  
We combine semantic alignment and structural coherence through a multiplicative score, which favors clusters that are simultaneously semantically aligned with the reference and structurally consistent with the seed region.
The final mask is obtained by merging the seed cluster with all candidate clusters whose combined score exceeds a similarity threshold~$\alpha$:
\begin{align}
    S_k &= s^{\text{cross}}_k  \cdot s^{\text{intra}}_k \label{eq:combined-score}\\
    \mathcal{M}_{\text{final}} &= \mathcal{G}^* \ \cup \ \bigl\{ \mathcal{G}_k \in \mathcal{C}_{\text{cand}} \,\big|\, S_k \geq \alpha \bigr\}.
\end{align}

\section{Experiments}
\label{sec:experiments}
We evaluate \ours{} on one-shot \emph{semantic}, \emph{part}, and \emph{personalized} segmentation.
In each setting, a single annotated reference mask is provided, and the model is tasked with segmenting the corresponding concept in the target image:
\emph{(1) semantic} -- all instances of a given class (\eg, “dog”);
\emph{(2) part} -- same object part (\eg, “dog ear”); \\
\emph{(3)~personalized} -- same object instance (\eg, “my dog”).

For \textbf{one-shot \emph{semantic} segmentation}, we use six datasets across a range of imaging scenarios: COCO-20$^i$~\cite{Nguyen:2019:Feature} with 80 object categories; LVIS-92$^i$~\cite{Liu:2023:Matcher} with 920 categories and a strong long-tail distribution; 
ISIC2018~\cite{Codella:2019:Skin, Tschandl:2018:Ham10000} for skin lesion segmentation; Chest X-Ray~\cite{Candemir:2013:Lung, Jaeger:2013:Automatic}, an X-ray dataset of lung screening; iSAID-5$^i$~\cite{Yao:2021:Isaid}, a remote sensing dataset with 15 categories; and SUIM~\cite{Islam:2020:Suim} with underwater imagery and 8 categories. For \textbf{one-shot \emph{part} segmentation}, we use PASCAL-Part~\cite{Liu:2023:Matcher}, providing 56 object parts across 15 categories, and PACO-Part~\cite{Liu:2023:Matcher} with 303 object parts from 75 categories. For \textbf{one-shot \emph{personalized} segmentation}, we use PerMIS~\cite{Samuel:2024:Waldo}, covering 16 categories.

\myparagraph{Implementation details.}
We adopt the Large version of the DINOv3~\cite{Simeoni:2025:Dinov3} encoder. 
Input images are resized to 1024 $\times$ 1024, following SAM-based approaches~\cite{Liu:2023:Matcher, Zhang:2023:PerSAM, Zhang:2024:GF-SAM}. The final segmentation masks are predicted at patch resolution: we bilinearly interpolate them to original resolution, and apply mask refinement with a CRF \cite{Krahebuhl:2011:CRF}, following \cite{Hamilton:2022:USS, Hahn:2025:CUPS, Gansbeke:2021:Unsupervised, Melas:2022:Spectral}.
We employ agglomerative clustering~\cite{Mullner:2011:Agglomerative} with $\tau=0.6$ and set the cluster aggregation threshold to $\alpha = 0.2$.
See Supplementary Material for more details.

\subsection{Main results}

\paragraph{Baselines.}
\Cref{tab:main_comparison} compares against state-of-the-art ICS baselines in terms of mean Intersection-over-Union (mIoU). 
The primary baselines are \emph{training-free methods}, specifically 
{PerSAM}~\cite{Zhang:2023:PerSAM}, {Matcher}~\cite{Liu:2023:Matcher}, and {GF-SAM}~\cite{Zhang:2024:GF-SAM}.
For GF-SAM, the strongest training-free baseline, we also include a variant in which DINOv2 is replaced with DINOv3 and a version with our feature debiasing to ensure a fair comparison. 
We also report \emph{task-specific fine-tuning} methods such as SegIC~\cite{Meng:2024:SEGiC} and DiffewS~\cite{Zhu:2024:Unleashing}, which leverage semantic and mask supervision. For SegIC, we also report a version trained only on COCO. While these operate under a different supervision regime, they provide an upper reference point for in-domain accuracy. We emphasize that \textbf{\ours} is the only method in \cref{tab:main_comparison} that uses no supervision (neither during pre-training nor fine-tuning) and operates solely on the self-supervised DINOv3 backbone.

\begin{figure*}
        \centering
    \includegraphics[width=\linewidth]{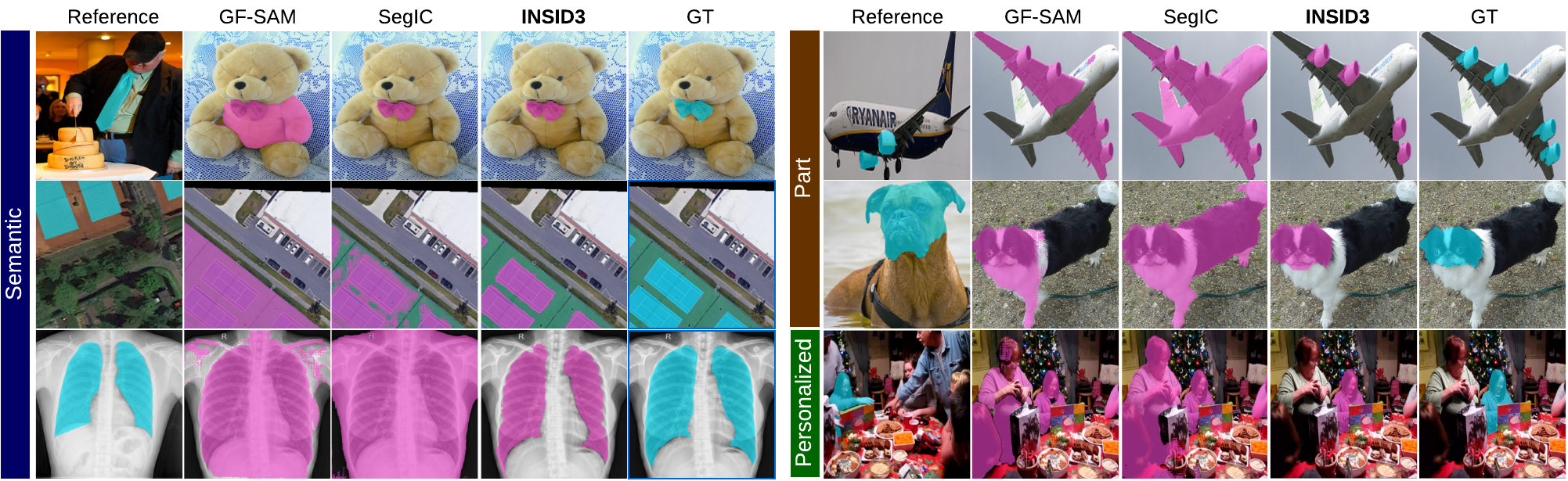}
    \vspace{-1.75em}
    \caption{Comparison of \textbf{\ours{}} with GF-SAM \cite{Zhang:2024:GF-SAM} and SegIC \cite{Meng:2024:SEGiC} on one-shot semantic \emph{(left)}, part \emph{(top right)}, and personalized \emph{(bottom right)} segmentation. \textbf{SegIC} performs well in-domain but struggles to generalize across domains and part granularity, reflecting its limited flexibility beyond the training distribution. \textbf{GF-SAM}, relying on the strong segmentation priors of SAM \cite{Kirillov:2023:SAM}, produces high-quality masks; however, the decoupled mechanism between correspondence and segmentation often leads to over- or under-segmentation. \textbf{\ours{}}, despite relying solely on self-supervised features, achieves precise localization and competitive mask quality.
    }
    \label{fig:qualitatives}
    \vspace{-0.5em}
\end{figure*}

\myparagraph{One-shot semantic segmentation.}
\Cref{tab:main_comparison} shows that \ours{} consistently outperforms training-free SAM-based pipelines, with gains over GF-SAM of \SI{+6.6}{\%} pts.\ mIoU on LVIS-92$^i$, \SI{+5.7}{\%} pts.\ on ISIC, \SI{+1.8}{\%} pts.\ on SUIM, and \SI{+27.8}{\%} pts.\ on Chest X-ray, \etc. 
Upgrading GF-SAM from DINOv2 to DINOv3 yields comparable performance, as its method relies only on sparse matched points to prompt SAM, thus discarding most of the information in DINOv3 dense features.
However, our debiasing helps GF-SAM with DINOv3 ( pts.\ mIoU on average).
In contrast, \ours{} performs estimation and segmentation in a unified space, achieving both higher accuracy and lower architectural complexity (\SI{304}{M} \vs \SI{945}{M} parameters).
Interestingly, fine-tuned methods achieve strong in-domain results (\eg, SegIC reaches \SI{76.1}{\%} mIoU on COCO-20$^i$) but drop sharply on other datasets, reflecting the inherent trade-off of specialization (\eg \SI{-6}{\%} \wrt \ours{} on iSAID).
Qualitative results in \cref{fig:qualitatives} \emph{(left)} show that \ours{} produces surprisingly clean segmentation masks directly from DINOv3 features, without any decoder or task-specific supervision.

\myparagraph{One-shot part segmentation.}
\ours{} achieves significant improvements over existing baselines on also on part segmentation (\cf \cref{tab:main_comparison}). It outperforms GF-SAM by \SI{+6.0}{\%} pts.\ mIoU on PASCAL-Part and \SI{+2.4}{\%} pts.\ mIoU on PACO-Part. Two-stage pipelines frequently over- or under-cover object parts due to fixed mask priors, inherited through fine-tuning or from SAM. In contrast, \ours{} better exploits the reference signal throughout the pipeline, enabling flexible and accurate part-level predictions. Compared to fine-tuned approaches, \ours{} outperforms SegIC and DiffewS by \SI{+10.6}{\%} / \SI{+16.5}{\%} pts. on PASCAL-Part and \SI{+12.8}{\%} / \SI{+15.9}{\%} pts. on PACO-Part. As illustrated in \cref{fig:qualitatives} \emph{(top right)}, \ours{} produces part masks that better preserve object structure and segmentation granularity.

\myparagraph{One-shot personalized segmentation.}
\ours{} achieves the best results on PerMIS, reaching \SI{67.0}{\%} mIoU and surpassing GF-SAM by \SI{+12.9}{\%} pts., SegIC by \SI{+15.2}{\%} pts., and DiffewS by \SI{+31.8}{\%} pts. This task is particularly challenging due to the presence of multiple visually similar distractor instances. Unlike previous work, which relies solely on positive activations from the reference and tends to segment all semantically related objects, \ours{} additionally exploits negative evidence through backward correspondences (\cref{subsec:matching}) to suppress irrelevant regions. \Cref{fig:qualitatives} \emph{(bottom right)} shows that this leads to accurate instance selection, even in the presence of visual ambiguity.

\definecolor{lightblue}{RGB}{0,114,200} 

\newcommand{\errlabel}[2]{\(\delta = \textcolor{red}{#1}\,/\;\textcolor{lightblue}{#2}\)} \newcommand{\imgwitherr}[3]{\begin{tikzpicture} \node[inner sep=0] (img) {\includegraphics[width=\linewidth]{#1}}; \node[ anchor=south, inner sep=0pt ] at ($(img.north west)!0.75!(img.north east)+(0pt,0pt)$) {\scriptsize \errlabel{#2}{#3}}; \end{tikzpicture} }

\begin{table}[t]
\centering
\small
\setlength{\tabcolsep}{1.75pt}
\caption{\textbf{Semantic correspondence on SPair-71k} (PCK@$T$ in \%, $\uparrow$). Comparison across DINOv3 backbones, w/ and w/o debiasing.}
\label{tab:debias_spair}
\vspace{-0.5em}
\begin{tabularx}{\linewidth}{@{} S[table-format=1.2] XS[table-format=2.1]S[table-format=2.1] @{\quad} S[table-format=2.1]S[table-format=2.1] @{\quad} S[table-format=2.1]S[table-format=2.1] @{}}
\toprule
&& \multicolumn{2}{c@{\quad}}{\textbf{Small}} & \multicolumn{2}{c@{\quad}}{\textbf{Base}} & \multicolumn{2}{c@{}}{\textbf{Large}} \\
\cmidrule(lr{1.5em}){3-4}\cmidrule(lr{1.5em}){5-6}\cmidrule(lr){7-8}
{$T$} && {\textit{original}} & {\textit{debias}} & {\textit{original}} & {\textit{debias}} & {\textit{original}} & {\textit{debias}} \\
\midrule
\bfseries 0.05 && 26.8 & \bfseries 27.9 & 29.2 & \bfseries 32.3 & 32.7 & \bfseries 33.6 \\
\bfseries 0.10 && 43.8 & \bfseries 45.7 & 45.0 & \bfseries 50.0 & 50.6 & \bfseries 52.0 \\
\bfseries 0.15 && 53.2 & \bfseries 55.6 & 54.0 & \bfseries 59.9 & 60.3 & \bfseries 62.1 \\
\bfseries 0.20 && 59.8 & \bfseries 62.6 & 59.8 & \bfseries 66.4 & 66.4 & \bfseries 68.6 \\
\bottomrule
\end{tabularx}
\vspace{-0.65em}
\end{table}

\begin{figure}[t]
\centering

\begin{minipage}[t]{0.49\linewidth}
\imgwitherr{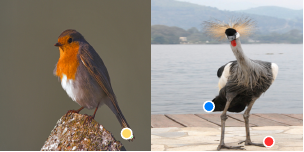}{\SI{46}{\%}}{\SI{8}{\%}}
\end{minipage}\hfill
\begin{minipage}[t]{0.49\linewidth}
\imgwitherr{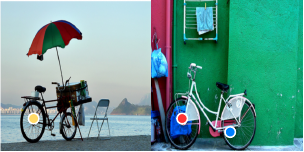}{61\%}{5\%}
\end{minipage}

\vspace{-5pt}

\begin{minipage}[t]{0.49\linewidth}
\imgwitherr{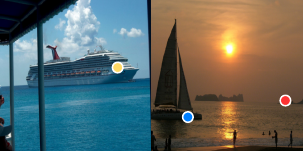}{\SI{97}{\%}}{\SI{3}{\%}}
\end{minipage}\hfill
\begin{minipage}[t]{0.49\linewidth}
\imgwitherr{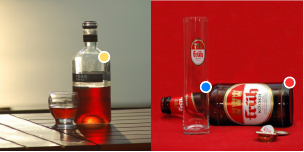}{\SI{57}{\%}}{\SI{7}{\%}}
\end{minipage}

\vspace{-5pt}
\tikz[baseline=-0.6ex]\draw[fill=yellow,draw=black] (0,0) circle (2.1pt);~\small{source keypoint;}\;
\tikz[baseline=-0.6ex]\draw[fill=red,draw=black] (0,0) circle (2.1pt);~prediction (original DINOv3);\;
\tikz[baseline=-0.6ex]\draw[fill=lightblue,draw=black] (0,0) circle (2.1pt);~prediction (debiased).\;
\(\delta\) is the relative error \wrt GT.
\vspace{-0.5em}
\caption{\textbf{Qualitative examples on SPair-71k} with DINOv3-L.}

\label{fig:debias_spair_qual}
\vspace{-0.5em}
\end{figure}

\subsection{On the positional bias of DINOv3 features}
\label{subsec:positional_bias}

\begin{figure}[t]
    \centering
    \pgfplotsset{
  compat=1.18,
  every axis/.append style={
    tick label style={font=\footnotesize},
    label style={font=\footnotesize},
    line width=0.6pt,
    mark options={fill=none, line width=0.4pt},
    mark size=1.2pt
  }
}
\tikzset{
  coco/.style   ={color=semantic, mark=*, mark options={fill=semantic}},
  paco/.style   ={color=part, mark=square*, mark options={fill=part}},
}
\tikzset{every picture/.style={/utils/exec={\sffamily\fontsize{7}{2}}}}
\newlength{\plotGap}
\setlength{\plotW}{3.2cm}   
\setlength{\plotGap}{-0.5mm}   
\setlength{\plotH}{3cm}   
\begin{tikzpicture}[>={Stealth[inset=0pt,length=4.5pt,angle'=45]}]
  \begin{axis}[
    enlargelimits=false,
    clip=true,
    name=leftax,
    width=\plotW, height=\plotH,
    scale only axis=true,
    enlarge x limits=false, enlarge y limits=false,
    xmin=0, xmax=1024, ymin=46.5, ymax=58,
    xtick={0,200,400,600,800,1024},
    xticklabels={0,200,400,600,800,1024},     
    ytick={48, 50, 52, 54, 56},
    yticklabels={48, 50, 52, 54, 56},
    xlabel={debiasing rank $s$}, ylabel={},
    ylabel={mIoU (in \%, $\uparrow$)},
    ylabel style={yshift=-2pt},
    xtick pos=bottom,
    ytick pos=left,
  ]
    \addplot+[coco] coordinates {(0,54.5)(100,56.90) (200,57.05) (300,57.14) (400,57.41) (500,57.65) (600,57.26) (700,57.05)(800,56.43)(900,55.44) };\label{pgfplots:coco};
    \addplot+[paco] coordinates {(0,47.8)(100,50.76) (200,50.79) (300,50.56) (400,50.42) (500,50.47) (600,50.41) (700,49.77) (800,49.65) (900,48.46)  };\label{pgfplots:paco};
     \draw[<-] (axis cs:30,54.5) -- (axis cs:130,54.5) node[right] {no debias\vphantom{g}};
     \draw[<-] (axis cs:30,47.8) -- (axis cs:130,47.8) node[right] {no debias\vphantom{g}};
  \end{axis}

  \begin{axis}[
    axis line style={-{Stealth[inset=0pt,length=4.0pt,angle'=45]}},
    enlargelimits=false,
    clip=true,
    at={(leftax.right of south east)},
    anchor=left of south west,
    xshift=\plotGap,
    width=\plotW, height=\plotH,
    scale only axis=true,
    enlarge x limits=false, enlarge y limits=false,
    xmin=0.40, xmax=0.925, ymin=34.5, ymax=59,
    xtick={0.40,0.50,0.60,0.70,0.80,0.9},
    xticklabels={0.4,0.5,0.6,0.7,0.8,0.9},
    ytick={35,40,45,50,55},
    yticklabels={35,40,45,50,55},
    xlabel={clustering granularity $\tau$},
    xtick pos=bottom,
    ytick pos=left,
  ]
    \addplot+[coco] coordinates {(0.40,45.48) (0.45,51.90) (0.55,56.54) (0.6,57.6) (0.65,56.78) (0.70,56.33) (0.75,55.99)(0.8, 55.5) (0.9,54.9)};
    \addplot+[paco] coordinates {(0.40,37.41) (0.45,41.10)(0.55,48.43) (0.6, 50.58) (0.65,51.74)(0.70,52.30) (0.75,52.58) (0.80,52.47) (0.9, 51.4)};
    \addplot[color=white, only marks, mark=star, mark size=2.0pt] coordinates {(0.60,59.6) (0.60,40.9) (0.60,54.5)};

     \draw[<-] (axis cs:0.425,37.25) -- (axis cs:0.478,37.25) node[right] {under-cl.};
     \draw[<-] (axis cs:0.895,37.25) -- (axis cs:0.842,37.25) node[left] {over-cl.};
    
    \end{axis}

    \node[draw,fill=white, inner sep=0.5pt, anchor=center] at (0.195\textwidth, 3.275) {\renewcommand{\arraystretch}{0.875}\setlength{\tabcolsep}{1.0pt}\fontsize{7}{2}
    \begin{tabular}{
    p{1.23cm}p{0.8cm}p{1.7cm}p{0.8cm}}
    \sffamily COCO-20$^i$\vphantom{g$^{a^a}$} & \ref*{pgfplots:coco} & \sffamily PASCAL-Part\hphantom{g} & \ref*{pgfplots:paco} 
    \end{tabular}};
    \end{tikzpicture}
    \vspace{-5pt}
    \caption{\textbf{\ours{} hyperparameters.} We study the effect of the debiasing rank $s$  \emph{(left)} and the clustering granularity $\tau$ \emph{(right)} on semantic (COCO-20$^i$) and part (PASCAL-Part) segmentation.}
     \label{fig:analysisparameters}
\end{figure}

While we focus on ICS, the positional bias we uncover in DINOv3 has broader implications for tasks that rely on semantic image alignment without spatial priors. A representative example is \emph{semantic correspondence} \cite{Zhang:2024:Telling, Tang:2023:Dift, Zhang:2023:Tale}, which evaluates how well dense features can localize semantically corresponding points across different images.
We evaluate our debiasing strategy on SPair-71k~\cite{Min:2019:Spair}, the standard benchmark for this task. Following common evaluation protocol~\cite{Tang:2023:Dift,Zhang:2023:Tale}, for each source keypoint, we compute cosine similarity to all target patch tokens and select the most similar one. Accuracy is measured using PCK@$T$ (\% of Correct Keypoints within a normalized distance $T$).

Quantitatively, \cref{tab:debias_spair} shows that debiasing leads to consistent gains of {+0.9–6.6} {PCK} across all model sizes. These results show that our positional debiasing acts as a simple training-free correction that improves the reliability of DINOv3 features for tasks requiring semantic alignment across images. Qualitatively, \cref{fig:debias_spair_qual} compares predictions obtained with the original DINOv3 features \emph{(red)} and our debiased features \emph{(blue)}. Predictions based on original features are influenced by a mix of semantic and positional cues, resulting in systematic errors. In contrast, debiased features align closely with the correct semantic location, yielding more semantically grounded matches.

\subsection{Ablation study}
We conduct ablations on COCO-20$^i$ and PASCAL-Part.
For more analyses, please refer to the Supplementary Material. 

\myparagraph{Debiased feature space.} 
We study the influence of the \emph{rank} $s$ of the estimated positional subspace $\mathbf{B}$ on our proposed \textit{debiasing} of DINOv3 from \cref{eq:debiasing}. 
\cref{fig:analysisparameters} \emph{(left)} shows a stable trend, improving over the \textit{no debiasing} baseline (\ie, $s=\text{\num{0}}$), indicating that the removed positional subspace does not carry semantically meaningful information.
Beyond an intermediate range, the gain saturates and eventually reverses once too many subspace dimensions are being removed. We fix $s$ to \num{500} across all datasets, which yields consistent gains (\SI{+3.1}{\%} on COCO, \SI{+2.7}{\%} on PASCAL and up to \SI{+6.6}{\%} PCK on SPair-71k, \cf \cref{tab:debias_spair}).

\begin{table}[t]
\caption{\textbf{Effect of clustering and aggregation.}
We compare our approach with tuned baselines on COCO-20$^i$ and PASCAL-Part (mIoU \%, $\uparrow$). All thresholds (0.55) and clustering granularities ($\tau{=}\text{\num{0.5}}$, $\tau{=}\text{\num{0.6}}$) are tuned to optimal values for fair comparison.}
\label{tab:ablationcomponents}
\vspace{-0.5em}
\small
\setlength{\tabcolsep}{3pt}
\renewcommand{\arraystretch}{1.02}
\begin{tabularx}{\linewidth}{@{}Xcc@{}}
\toprule
\textbf{Variant} & \textbf{COCO} & \textbf{PASCAL-Part} \\
\midrule
\multicolumn{3}{@{}l}{\emph{No clustering}} \\
~~Thresholding sim.\ map @ 0.55 (Eq.~\ref{eq:naive_prototype}) & 44.2 & 35.4 \\
\cmidrule(lr){1-3}
\multicolumn{3}{@{}l}{\emph{Clustering w/o aggregation}} \\
~~Coarse clustering ($\tau=\text{\num{0.5}}$) & 50.6 & 31.1 \\
~~Fine clustering ($\tau=\text{\num{0.6}}$) & 42.8 & 36.2
 \\
\cmidrule(lr){1-3}
\multicolumn{3}{@{}l}{\textbf{Ours}: \emph{Clustering w/ aggregation} ($\tau=\text{\num{0.6}}$)} \\
~~w/ cross similarity & 54.6 & 48.5 \\
~~w/ self + cross similarity (Eq.~\ref{eq:combined-score}) & \textbf{57.6} & \textbf{50.5} \\ 
\bottomrule
\end{tabularx}
\vspace{-0.5em}
\end{table}

\myparagraph{Cluster granularity.}
Next, we analyze the impact of the \emph{similarity threshold} $\tau$ of agglomerative clustering (\cref{sec:clustering}). 
\cref{fig:analysisparameters} \emph{(right)} shows that finer partitions resulting from larger $\tau$ are beneficial for part-level tasks. Conversely, over-clustering is detrimental to object-level tasks, fragmenting coherent objects into many small clusters. We set $\tau = \text{\num{0.6}}$ across tasks, providing a sensible trade-off between part sensitivity and semantic coherence.

\myparagraph{Clustering and aggregation.}
We lastly analyze the role of clustering and aggregation in \cref{tab:ablationcomponents}.
Thresholding the similarity map (Eq.~\ref{eq:naive_prototype}) provides a coarse baseline, yielding \SI{44.2}{\%} and \SI{35.4}{\%} mIoU on COCO and PASCAL-Part.
Introducing clustering without aggregation, where a single cluster is selected, improves results but requires tuning the granularity parameter $\tau$ separately for each task: lower values ($\tau = \text{\num{0.5}}$) better capture full objects, while higher ones ($\tau = \text{\num{0.6}}$) suit part-level structures.
Even tuned independently, this variant still underperforms.
Our solution fixes an intermediate $\tau$ and resolves this trade-off through a principled aggregation strategy (\cf \cref{subsec:aggregation}).
Aggregating by cross-image similarity already improves segmentation results, achieving \SI{54.6}{\%} and \SI{48.5}{\%}, while jointly leveraging cross- and self-similarity yields the best results, reaching \SI{57.6}{\%} and \SI{50.5}{\%} mIoU on COCO and PASCAL-Part.

\section{Conclusion}
\label{sec:conclusion}
In this work, we introduced \ours{}, a training-free framework for in-context segmentation built solely on DINOv3. By leveraging the dual nature of DINOv3 features, \ie semantic alignment and spatial coherence, \ours{} performs correspondence estimation and segmentation within a single backbone. Despite its simplicity, it exhibits strong generalization across one-shot \textit{semantic}, \textit{part}, and \textit{personalized segmentation}, outperforming both fine-tuned and training-free approaches. Overall, this suggests that segmentation can emerge naturally from self-supervised dense representations. While existing methods rely on either fine-tuning or training-free pipelines grounded in mask-supervised pre-training, \ours{} remains fully \textit{unsupervised}, relying solely on the in-context example for guidance.
This suggests that reducing supervision may foster more robust and transferable representations, marking a concrete step toward more scalable and general-purpose visual understanding.

{\small \inparagraph{Acknowledgments.} 
Claudia Cuttano was supported by the Sustainable Mobility Center (CNMS), which received funding from the European Union Next Generation EU (Piano Nazionale di Ripresa e Resilienza (PNRR), Missione 4 Componente 2 Investimento 1.4 ``Potenziamento strutture di ricerca e creazione di `campioni nazionali di R\&S' su alcune Key Enabling Technologies'') with grant agreement no.\ CN\_00000023. Christoph Reich is supported by the Konrad Zuse School of Excellence in Learning and Intelligent Systems (\href{https://eliza.school}{ELIZA}) through the DAAD programme Konrad Zuse Schools of Excellence in Artificial Intelligence, sponsored by the German Federal Ministry of Education and Research. 
Stefan Roth has received funding from the European Research Council (ERC) under the European Union's Horizon 2020 research and innovation programme (grant agreement No.\ 866008).
Further, he was supported by the DFG under Germany's Excellence Strategy (EXC-3057/1 ``Reasonable Artificial Intelligence'', Project No.~533677015) and by the LOEWE initiative (Hesse, Germany) within the emergenCITY center [LOEWE/1/12/519/03/05.001(0016)/72]. Daniel Cremers has received funding by the European Research Council (ERC) Advanced Grant SIMULACRON (grant agreement No.\ 884679). We acknowledge the CINECA award
under the ISCRA initiative, for the availability of high-performance computing resources.
We also acknowledge the support of the European Laboratory for Learning and Intelligent Systems (ELLIS). Finally, we thank Bar{\i}{\c{s}} Z{\"o}ng{\"u}r for the insightful feedback.}

{
    \small
    \bibliographystyle{ieeenat_fullname}
    \bibliography{bibtex/short, bibtex/references}
}

\clearpage
\setcounter{section}{0}
\renewcommand\thesection{\Alph{section}}
\setcounter{page}{1}
\pagenumbering{roman}
\twocolumn[{%
\renewcommand\twocolumn[1][]{#1}%
\maketitlesupplementary
{\large Claudia Cuttano\textsuperscript{\normalfont{}* 1,2}
\authorstep Gabriele Trivigno\textsuperscript{\normalfont{}* 1}
\authorstep Christoph Reich\textsuperscript{\normalfont{}\,2,3,5,6}\\
Daniel Cremers\textsuperscript{\normalfont{}\,3,5,6}
\authorstep Carlo Masone\textsuperscript{\normalfont{}\,1}
\authorstep Stefan Roth\textsuperscript{\normalfont{}\,2,4,5}\\[2.75pt]
\small{\textsuperscript{1}Politecnico di Torino\affiliationstep \textsuperscript{2}TU Darmstadt\affiliationstep \textsuperscript{3}TU Munich\affiliationstep \textsuperscript{4}hessian.AI\affiliationstep \textsuperscript{5}ELIZA\affiliationstep \textsuperscript{6}MCML\affiliationstep
\textsuperscript{*}equal contribution}\\[-5pt]\small {\url{https://visinf.github.io/INSID3}}
\vspace{0.75cm}
}
}]

\noindent In this appendix, we provide additional analyses, implementation details, and experiments for \ours{}. 

\noindent Specifically:
\begin{itemize}
    \item \textbf{Positional bias}. In \cref{sec:positional}, we further analyze the issue of positional bias in DINOv3, including comparisons with DINOv2 and additional debiasing studies.
    \item \textbf{Implementation details}. In \cref{sec:hyperpars}, we report the hyperparameters used throughout the paper. The same set of hyperparameters is fixed across all datasets and tasks.
    \item \textbf{Additional experiments}. In \cref{sec:add_exps}, we present further experiments, including the 5-shot setting, backbone comparisons, evaluation against SAM 3, and the empty-mask corner case.
    \item \textbf{Computational cost}. In \cref{sec:computational}, we analyze the computational cost of our method and compare it against training-free and fine-tuned baselines.
    \item \textbf{Qualitative examples}. In \cref{sec:qualitatives}, we report additional qualitative comparisons and examples of our method.
    \item \textbf{Limitations and future directions}. In \cref{sec:limitations}, we discuss current limitations of \ours{} and outline possible extensions.
\end{itemize}

\begin{figure*}[b]
\centering
\includegraphics[width=\linewidth]{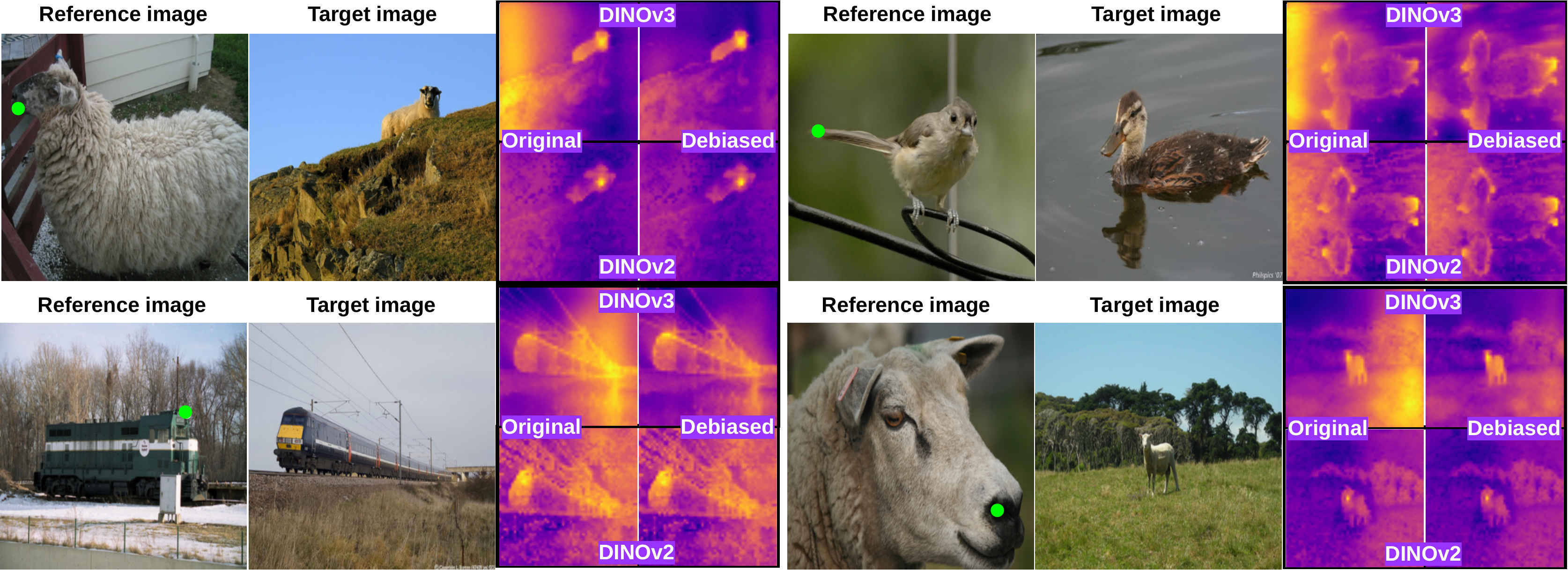}
\vspace{-1.75em}
\caption{\textbf{Positional bias in DINOv2 \vs DINOv3.}
We visualize the similarity map between a reference keypoint and all target patches using \cref{eq:supp_sim_map}.  
Darker purple regions indicate low similarity, while brighter yellow regions indicate high similarity.  
DINOv3 exhibits strong coordinate-aligned artifacts, with bright responses appearing at the same absolute spatial location as the reference keypoint. Our debiasing projection suppresses these spurious activations.  
We observe that DINOv2 shows much weaker positional bias.}

\label{fig:qualitatives_dinov2}
\vspace{-0.5em}
\end{figure*}

\section{On the Positional Bias}
\label{sec:positional}

\subsection{DINOv2 vs. DINOv3: Positional bias}
\label{sec:supp_positional_bias}

To assess how strongly positional bias is encoded in DINOv3 \cite{Simeoni:2025:Dinov3} versus
DINOv2 \cite{Oquab:2023:Dinov2}, we analyze the similarity maps from both encoders, computed as in \cref{sec:unlocking} of the main paper. Given a
reference image with one annotated keypoint and a target image, we extract
dense patch embeddings $\mathbf{F}^r$ and $\mathbf{F}^t$ from the respective backbone, let $j$
denote the patch index of the keypoint, and form the reference prototype
\begin{equation}
\mathbf{p}^r=\mathbf{F}^{r}_{j}.
\end{equation}
We then compute the target similarity map as
\begin{equation}
\label{eq:supp_sim_map}
\mathrm{sim}(i)=\langle \mathbf{F}^t_i,\, \mathbf{p}^r\rangle,\qquad i\in\{1,2,\dots,P\}.
\end{equation}
For each backbone, we visualize \emph{(i)} the similarity map obtained from the
original features, and \emph{(ii)} the similarity map obtained after applying our debiasing
projection from \cref{eq:noise_feats,eq:debiasing} from the main paper. These similarities are visualized in \cref{fig:qualitatives_dinov2}. The maps show a notable difference between the two models. With the original
\emph{DINOv3} features, we observe pronounced coordinate-aligned artifacts: patches located at the same absolute position as the reference keypoint consistently exhibit strong responses.
This phenomenon is more prominent in background regions, where semantic content is scarce or absent, exposing the positional component, which then emerges as spurious activations. After applying the debiasing projection, these position-driven responses are suppressed while the similarity patterns on the object remain stable. This shows that our debiasing suppresses positional bias without degrading the semantic structure encoded in the features.
\textbf{DINOv2}, in comparison, exhibits weaker positional structure even without any debiasing. The coordinate-aligned patterns characteristic of DINOv3 are mostly absent, and debiasing leads to minor modifications. This contrast suggests that DINOv3 encodes more positional information. We suspect that this could arise as a by-product of its training objective, where the local-consistency constraints and Gram-anchoring may inadvertently amplify absolute spatial correlations in regions with weak semantic cues. The different positional encoding strategy used in DINOv3 compared to DINOv2 may further contribute to this effect.

\begin{table}[t]
\centering
\small
\caption{\textbf{Semantic correspondence on SPair-71k.} 
PCK@$T$ (\%, $\uparrow$) for DINOv2 and DINOv3 using original and debiased 
features. Debiasing produces marginal improvements for DINOv2, but yields 
substantial gains for DINOv3.}
\label{tab:debias_spair_supp}
\vspace{-0.4em}
\begin{tabularx}{\columnwidth}{@{}XXccc@{}}
\toprule
\textbf{Backbone} & \textbf{Features} & \textbf{PCK@0.05} & \textbf{PCK@0.10} & \textbf{PCK@0.20} \\
\midrule

\multirow{2}{*}{\textbf{DINOv2}}
  & original & 34.4 & 50.5 & 66.8 \\
  & debias   & \textbf{34.7} {\scriptsize(+0.3)} & \textbf{50.8} {\scriptsize(+0.3)} & \textbf{67.1} {\scriptsize(+0.3)} \\
\midrule

\multirow{2}{*}{\textbf{DINOv3}}
  & original & 29.2 & 45.0 & 59.8 \\
  & debias   & \textbf{32.3} {\scriptsize(+3.1)} 
             & \textbf{50.0} {\scriptsize(+5.0)} 
             & \textbf{66.4} {\scriptsize(+6.6)} \\
\bottomrule
\end{tabularx}
\vspace{-0.5em}
\end{table}

To quantitatively analyze this behavior, we evaluate semantic correspondence on
SPair-71k~\cite{Min:2019:Spair}, following the protocol described in \cref{subsec:positional_bias} of the main paper. 
This benchmark isolates the \textit{correspondence} component of the representation by measuring 
cross-image alignment directly on the backbone features, without the overhead of segmentation.
As shown in \cref{tab:debias_spair_supp}, debiasing yields only small gains for \textbf{DINOv2} (all improvements within
\SI{+0.3}{PCK}), consistent with its limited positional coupling. In sharp contrast, \textbf{DINOv3} benefits
substantially, with improvements of \SI{+3.1}{PCK} at $T = \num{0.05}$, \SI{+5.0}{PCK} at $T = \num{0.10}$, and
\SI{+6.6}{PCK} at $T = \num{0.20}$. These gains are significant at all thresholds and reflect the
removal of positional components that are detrimental for tasks involving feature matching across images. 

\subsection{Alternative debiasing strategies}
\label{sec:supp_alt_debias}

We compare our debiasing scheme against training-free alternatives on COCO-20$^i$, reported in \cref{tab:debis_ablation}. As a baseline, we propose to reduce positional sensitivity by averaging features over multiple augmented views of each image. Concretely, for each image, we sample $n \in \{4,12\}$ views using random horizontal and vertical flips as well as homographies. We feed each view independently to DINOv3 and average the corresponding patch features before performing matching and segmentation. This simple strategy already improves over the non-debiased baseline (from \SI{54.5}{\percent} to \SI{55.7}{\percent} and \SI{56.5}{\percent} mIoU for $n=\num{4}$ and $n=\num{12}$, respectively), but \emph{requires $n$ forward passes per each image} at inference time. In contrast, our projection-based debiasing (SVD from a single noise image) achieves a larger gain (\SI{57.6}{\percent} mIoU) while adding virtually no overhead beyond a single matrix multiplication at inference time. Importantly, our debiasing projection is computed once and stored offline. To test the stability of the estimated positional subspace, we alternatively perform SVD on the features from a small pool of images with \emph{low semantic content}: black, white, horizontal gradient, vertical gradient, and Gaussian noise patterns. Using 5 or 10 such images yields virtually identical results (\SI{57.7}{\percent} mIoU), confirming that the positional subspace is stable and can be captured reliably with a single noise realization.

\begin{table}[t]
\centering
\small
\setlength{\tabcolsep}{19pt}
\renewcommand{\arraystretch}{1.1}
\caption{\textbf{Different debiasing strategies.}
Comparison of training-free strategies for feature debiasing on COCO-20\textsuperscript{i} using mIoU (in \%, $\uparrow$). Augmentations are a viable alternative but require \emph{multiple forward passes}. In contrast our method adds no overhead, except for a single matrix multiplication, since the debiasing projection is pre-computed offline and stored.}
\vspace{-0.55em}
\begin{tabularx}{\columnwidth}{@{}Xc@{}}
    \toprule
    \textbf{Method} 
    & \textbf{mIoU} \\
    \midrule
    {No debiasing} & 54.5 \\
    {Augmentations, 4 views}  & 55.7 \\
    {Augmentations, 12 views}  &  56.5 \\
    {\textbf{Ours} (SVD from 1 noise image)}  & 57.6 \\
    \midrule
    SVD from 5 noise images & 57.7\\
    SVD from 10 noise images & 57.7 \\
    \bottomrule
    \end{tabularx}
\vspace{-0.6em}
\label{tab:debis_ablation}
\end{table}

\subsection{Semantics in debiased feature space}
\label{sec:supp_pca_debias}

To further examine whether our debiasing projection removes primarily positional variance while preserving the semantic information of the representation, we analyze how semantic information is distributed across principal components in the original and debiased feature spaces. The key idea is simple: in the original DINOv3 representation, positional components account for a substantial amount of variance, causing PCA to allocate some of the leading directions to positional rather than semantic structure. After applying the projection (\cf \cref{eq:debiasing} of the main paper), this positional subspace is removed, reducing the effective rank of the feature representation: although the channel dimensionality remains unchanged, the directions associated with positional variation are collapsed. Therefore, if debiasing removes positional directions, the remaining variance no longer reflects positional structure, and the resulting principal components predominantly encode semantic variability.

We test this hypothesis by measuring cross-image semantic correspondence accuracy as a function of the retained dimensionality $d$. The PCA of original features should degrade earlier when reducing $d$, since its components contain a mixture of semantic and positional signals. 
This experiment complements the analysis in \cref{fig:analysisparameters} of the main paper, which focuses on in-context segmentation, by directly probing the effect of debiasing on cross-image feature matching.

The experimental protocol is as follows. For both the \textit{original} and the \textit{debiased} feature spaces, we
compute PCA using a separate set of training images, obtaining two orthonormal
bases $\mathbf{U}_{\text{orig}}$ and $\mathbf{U}_{\text{debias}}$. This
allows us to study how semantic information is distributed when the feature
representation is constrained to $d$ dimensions. For each
$d$ we evaluate:

\begin{itemize}
    \item \textbf{PCA-original.}  
    We compute the PCA basis $\mathbf{U}_{\text{orig}}$ on original DINOv3 features extracted from COCO-20$^i$ training images. At test time, target features $\mathbf{F}$ are projected onto the top-$d$ components:
    \begin{equation}
      \mathbf{F}^{(d)}_{\text{orig}}
        = \mathbf{F}\,\mathbf{U}_{\text{orig}}[:,1{:}d].
    \end{equation}

    \item \textbf{Debias+PCA.}  
    We first apply our positional debiasing projection to the same COCO-20$^i$ training features, then compute a PCA basis $\mathbf{U}_{\text{debias}}$ in this debiased space. At test time, features are first debiased, then projected onto the top-$d$ PCA components:
    \begin{equation}
      \tilde{\mathbf{F}}^{(d)}_{\text{PCA}}
        = \tilde{\mathbf{F}}\,\mathbf{U}_{\text{debias}}[:,1{:}d].
    \end{equation}
\end{itemize}
Semantic correspondence is evaluated on SPair-71k, isolating cross-image alignment from
segmentation, by using these two
$d$-dimensional representations.
Results are plotted in \cref{fig:pca_vs_debias}. Across all dimensionalities, the PCA-debiased curve remains consistently above the PCA-original. This shows that the variance removed by our projection does not encode meaningful semantic information. Once positional variance is eliminated, the remaining dimensions form a cleaner and more stable semantic subspace. In contrast, the leading components of the \emph{original} features also capture structured positional signals, resulting in consistently lower correspondence accuracy.
In particular, at $d=\text{\num{32}}$, the plot shows a gap of \num{+7.8} \pckTen{}. As $d$ increases, both representations approach their full-dimensional capacity, and the gap converges to \num{+2.8} \pckTen{}. While this experiment does not establish that \emph{all} positional biases are removed, it provides clear evidence that the directions suppressed by our debiasing are not useful for identifying semantic correspondences, and that the resulting representation supports more reliable matching across a wide range of dimensionalities.

\begin{figure}[t]
    \centering
    \pgfplotsset{
      compat=1.18,
      every axis/.append style={
        tick label style={font=\footnotesize},
        label style={font=\footnotesize},
        line width=0.6pt,
        mark options={fill=none, line width=0.4pt},
        mark size=1.2pt
      }
    }

    \tikzset{
      pcaonly/.style   ={color=semantic, mark=*, mark options={fill=semantic}, dashed},
      debiaspca/.style ={color=part,     mark=square*, mark options={fill=part}},
    }
    \tikzset{every picture/.style={/utils/exec={\sffamily\fontsize{7}{2}}}}

    \setlength{\plotW}{7.1cm}   
    \setlength{\plotH}{4.25cm}   

    \begin{tikzpicture}[>={Stealth[inset=0pt,length=4.5pt,angle'=45]}]
      \begin{axis}[
        enlargelimits=false,
        clip=true,
        width=\plotW, height=\plotH,
        scale only axis=true,
        enlarge x limits=false, enlarge y limits=false,
        xmin=0, xmax=1024,
        ymin=20.5, ymax=54,
        xtick={30,100,200,400,600,800, 1000},
        xticklabels={30,100,200,400,600,800, 1000},
        ytick={22, 26, 30, 34, 38, 42,46,50,54,58},
        yticklabels={22, 26, 30, 34, 38, 42,46,50,54,58},
        xlabel={Dimensionality $d$},
        ylabel={PCK@0.10 (in \%, $\uparrow$)},
        ylabel style={yshift=-1pt},
        xtick pos=bottom,
        ytick pos=left,
      ]

        \addplot+[pcaonly] coordinates {
          (32,  21.4)
          (64,  32.4)
	       (96, 38.7)
          (128, 41.9)
          (256, 46.7)
          (384, 48.5)
          (512, 50.4) 
          (768, 50.8)
          (896, 51.1)
           (1000, 50.7)
        }; \label{pgfplots:pcaonly}

        \addplot+[debiaspca] coordinates {
        (32,  29.2)
        (64,  38.4)
	    (96, 42.5)
        (128, 45.1)
        (256, 48.9)
        (384, 50.0)
        (512, 51.8) 
        (768, 52.8)
        (896, 53.1)
        (1000, 53.5)
        }; \label{pgfplots:debiaspca}
      \end{axis}

      \node[draw,fill=white, inner sep=0.9pt, anchor=south east] at (rel axis cs:1,0.4) {%
        \renewcommand{\arraystretch}{0.9}\setlength{\tabcolsep}{1.0pt}\fontsize{8}{2}\selectfont
        \begin{tabular}{p{1.9cm}p{0.8cm}}
          \sffamily Debias + PCA & \ref*{pgfplots:debiaspca} \\[2pt]
          \sffamily PCA - original   & \ref*{pgfplots:pcaonly} 
        \end{tabular}
      };
    \end{tikzpicture}
    \vspace{-4pt}

\caption{\textbf{Debiased features preserve semantic structure.}
We compute PCA bases on COCO training images for \emph{(i)} the original DINOv3 features and \emph{(ii)} our debiased features. We then project both representations to $d$ principal components and evaluate semantic correspondence (\pckTen{}) on SPair-71k.
Across all $d$, Debias + PCA consistently outperforms PCA-original, indicating that the removed directions primarily encode positional bias. After debiasing, the remaining variance concentrates around semantic structure, yielding better results when compressed via PCA.}
\vspace{-0.5em}
\label{fig:pca_vs_debias}
\end{figure}

\begin{table}[t]
\centering
\small
\renewcommand{\arraystretch}{1.1}
\setlength{\tabcolsep}{4pt}
\caption{\textbf{Hyperparameter overview.} \ours{} uses only three hyperparameters, all fixed across datasets, for both semantic, part, and personalized segmentation.}
\vspace{-0.4em}
\begin{tabularx}{\columnwidth}{@{}lXc@{}}
\toprule
\multicolumn{2}{@{}l}{\textbf{Hyperparameter}} & \textbf{Value} \\
\midrule
$\tau$ & (clustering sensitivity) & 0.6 \\
$\alpha$ &  (aggregation threshold) & 0.2 \\
$s$ &  (debias rank) & 500 \\
\bottomrule
\end{tabularx}
\vspace{-0.5em}
\label{tab:hyperpars}
\end{table}

\subsection{Discussion}

We proposed a simple approach for removing positional biases from dense features of DINOv3. We demonstrate that debiased features aid accuracy for cross-image matching. Still, positional information remains a trade-off. While positional information can support approaching certain tasks, strong absolute positional biases can hamper cross-image correspondence. For example, in semantic segmentation, predicting sky can benefit from absolute positional information, whereas correspondence tasks require features that generalize across spatial positioning. Existing work in different domains also removed absolute positional information explicitly or implicitly from dense features by exploiting equivariance across augmented views~\cite{Yang:2024:DViT, Fu:2024:FeatUp, Wimmer:2025:AnyUp} or multi-view consistency~\cite{Yue:2024:FiT3D, Jevtic:2025:SceneDINO, Yang:2024:EmerNeRF}. Different from these approaches, our approach does not require any training and provides a flexible, efficient, and simple decomposition between semantic features and absolute positional encoding.

\section{Implementation details}
\label{sec:hyperpars}

\ours{} relies on only three scalar hyperparameters, summarized in \cref{tab:hyperpars}.  
All hyperparameters are selected once on the COCO-20$^i$ training split using a $k$-fold
cross-validation procedure (with $k=\text{\num{3}}$), and are kept \emph{fixed across all datasets, domains,
and tasks} (semantic, part, and personalized segmentation). No dataset-specific or task-specific
tuning is performed. The clustering sensitivity $\tau$ controls the granularity of the agglomerative clustering
step. We choose a value ($\tau = \text{\num{0.6}}$) that provides fine-grained clusters at part level, and rely on aggregation based on similarity with the in-context example to merge clusters to the desired granularity.
The aggregation threshold
$\alpha$ determines how strictly clusters must agree semantically and structurally with the seed
region; the debiasing rank $s$ specifies the number of positional directions removed by our
projection. Despite their simplicity, these three values generalize well across all experiments, highlighting
the stability of the method.

\begin{table*}[t]
\caption{\textbf{Comparison of \ours{} (mIoU in \%, $\uparrow$) on 5-shot semantic and part segmentation.} Models are provided with 5 contextual examples and tasked with segmenting the annotated concept in the target image. 
\ours{} scales effectively to multiple references, achieving robust performance across domains. 
All hyperparameters are reused from the 1-shot setting without any tuning, highlighting the versatility of our approach.
\textcolor{indomain}{Gray} indicates the model was trained on the corresponding train split of the dataset; best results \textbf{bold}, \nth{2} best \underline{underlined}.
$\dagger$ denotes a GF-SAM variant using DINOv3 features.
}

\label{tab:kshot}
\vspace{-5pt}
\small
\setlength\tabcolsep{2.8pt}
\begin{tabularx}{\linewidth}{@{}Xlc|cccccc|cc|c@{}}
\toprule
& &  & \multicolumn{6}{c}{\textbf{Semantic}} & \multicolumn{2}{c}{\textbf{Part}}  &  \\
\textbf{Method} & \textbf{Encoder} & \textbf{\#{}Param} & 
 \textbf{LVIS-92\textsuperscript{i}}  &
 \textbf{COCO-20\textsuperscript{i}} &
 \textbf{ISIC} &
 \textbf{SUIM} & 
 \textbf{iSAID} &
 \textbf{X-Ray} &
 \textbf{PASCAL} &
 \textbf{PACO} &
 \textbf{Avg}\\
\midrule
\multicolumn{4}{@{}l}{\textbf{Task-specific fine-tuning}: \textit{Semantic + mask supervision}} \\
~~SegGPT  \cite{Wang:2023:SegGPT} & \small{ViT} & \SI{354}{M} & 25.4 & \ind{67.9} & 45.2 & 33.7 & \ind{35.9} & \textbf{89.1} &  42.8 & \ind{14.1} & 44.3 \\ 
~~SINE   \cite{Liu:2024:SINE} & \small{DINOv2} & \SI{373}{M} & 35.5 & \ind{66.1} & 28.6  & 54.8 & 40.5 & 40.6 & 36.4 & 25.4  & 41.0 \\ 
~~DiffewS \cite{Zhu:2024:Unleashing} & \small{Stable Diffusion} & \SI{890}{M} & 35.4 & \ind{72.2} & 32.7 & 49.8 & 48.0 & 45.1 & 39.7 & 26.1  & 43.6 \\ 
\midrule
\multicolumn{4}{@{}l}{\textbf{Training free}: \textit{Mask-supervised pre-training}} \\
~~Matcher \cite{Liu:2023:Matcher} & \small{DINOv2 + SAM} & \SI{945}{M} & 40.0 & 60.7 & 35.0 & 50.6 & 34.3 & 71.2 & 45.8 & 33.6 & 46.4 \\ 
~~GF-SAM \cite{Zhang:2024:GF-SAM} & \small{DINOv2 + SAM} & \SI{945}{M} & \underline{44.2} & \textbf{66.8} & 55.2 & 58.1 & 52.4 & 52.9 & 51.2 & \underline{41.9}  & 52.8 \\ 
~~GF-SAM\textsuperscript{$\dagger$} \cite{Zhang:2024:GF-SAM} & \small{DINOv3 + SAM} & \SI{945}{M}  & 42.8 & 64.4 & 56.7 & 58.6 & 53.2 & 54.7 & 50.3 & 39.3 & 52.5\\ 
~~ \tikz[baseline=1ex]{\draw[->, thick] (0., 2.8ex) -- (0,1.4ex) -- (1em,1.4ex);} + \textit{our debias} & \small{DINOv3 + SAM} & \SI{945}{M} & 43.6 & 64.6 & \underline{58.2} & \underline{59.2}  & \underline{54.1} & 59.1 & \underline{51.4} & 40.2 & \underline{53.8}\\ 
\midrule
\multicolumn{11}{@{}l}{\textbf{Training free}: \textit{Unsupervised pre-training}} \\
~~\textbf{\ours} \emph{(ours)}   & {DINOv3}       & \textbf{\SI{304}{M}} & \textbf{47.2} & \underline{65.1} & \textbf{63.9} & \textbf{61.7} & \textbf{56.9} & \underline{80.1} & \textbf{57.1} & \textbf{46.8} & \textbf{59.9} \\
\bottomrule
\end{tabularx}
\vspace{-0.1em}
\end{table*}

\section{Additional Experiments}
\label{sec:add_exps}
Here, we provide additional experimental results, complementing \cref{sec:experiments} of the main paper.

\subsection{$k$-shot segmentation}
\label{subsec:kshot}

Although our main experiments evaluate INSID3 under the 1-shot segmentation setting, our approach naturally extends to multiple reference examples.  Let $\{(\mathbf{I}^{r_m}, \mathbf{M}^{r_m})\}_{m=1}^{k}$ denote the $k$ reference images and masks of the $k$-shot setting. The clustering of the target image (\cref{sec:clustering}) and the aggregation procedure (\cref{subsec:aggregation}) remain unchanged. Only the correspondence stage (\cref{subsec:matching}) requires adaptation.

For each target patch $i$, we compute its nearest neighbor in every reference image using the debiased features:
\begin{equation}
\text{NN}_m(i)=\argmax_{j\in\Omega}\,\langle \tilde{\mathbf{F}}^{t}_i,\tilde{\mathbf{F}}^{r_m}_j\rangle .
\tag{15}
\end{equation}
A patch is retained as a valid candidate if its nearest neighbor in the reference lies within the mask for a majority of the reference images, \ie,

\begin{equation}
\Bigl|\{\,m \mid \mathbf{M}^{r_m}_{\text{NN}_m(i)}=1\,\}\Bigr| \; \geq \; \Biggl\lceil \frac{k}{2} \Biggr\rceil .
\end{equation}
This majority-vote filtering keeps only correspondences consistently supported across references. For prototype construction, we compute one debiased prototype per reference, and then average them:
\begin{equation}
\tilde{\mathbf{p}}^{r} = \frac{1}{k}\sum_{m=1}^{k}\tilde{\mathbf{p}}^{r_m}.
\end{equation}
The aggregated prototype is plugged into \cref{eq:cross-similarity} to compute cross-image similarity for each target cluster, which drives seed selection and subsequent aggregation.
We follow standard few-shot segmentation protocols and evaluate the case $k=5$ in \cref{tab:kshot}.

\myparagraph{Discussion.}
We observe that \ours{} benefits consistently from additional reference examples. Moving from the 1-shot to the 5-shot setting, our method yields substantial gains across all benchmarks (\cf Tabs.\ \ref{tab:main_comparison} \& \ref{tab:kshot}), with improvements in the range of +\num{1.3}–\SI{9.5}{\percent} pts.\ mIoU. Overall, \ours{} achieves the best average score with a gain over the best competitor of \SI{+6.1}{\%} points. In particular, \ours{} outperforms GF-SAM by \SI{+3.0}{\%} pts.\ mIoU on the challenging LVIS-92$^i$, and by \SI{5.4}{\%} pts.\ on average on Part segmentation. On challenging datasets from the medical domain, the gain is \SI{+8.7}{\%} and \SI{+27.2}{\%} pts.\ on ISIC and X-Ray, respectively. On SUIM the performance boost is \SI{+3.6}{\%} pts., and \SI{+4.5}{\%} pts.\ mIoU on the remote-sensing domain of iSAID. Only on COCO, \ours{} incurs in a small gap of \SI{-1.7}{\%} pts.
Notably, our proposed debiasing strategy also improves the baseline of GF-SAM where DINOv3 is used as encoder, providing an improvement of \SI{+1.0}{\%} pts.\ over the original method with DINOv2.
Overall, the table reveals a trend consistent with the 1-shot setting: fine-tuned methods struggle to generalize when the test distribution deviates from the training distribution. Even when excluding the challenging medical and remote-sensing datasets, models fine-tuned on COCO exhibit a substantial accuracy drop on LVIS, whose long-tailed taxonomy differs from that of COCO. In contrast, training-free approaches avoid these pitfalls, offering stronger generalization across tasks and datasets without requiring domain-specific adaptation. Among them, our method stands out by leveraging only self-supervised DINOv3 features, yet achieving results that are competitive with or superior to methods relying on explicit mask-level supervision. Importantly, these 5-shot results are obtained \emph{without} introducing any new components or tuning additional hyperparameters: we reuse exactly the same encoder, clustering configuration, and aggregation thresholds as in the 1-shot experiments, and only replace the correspondence stage with the simple majority-vote and prototype averaging scheme in \cref{subsec:kshot}. This shows that \ours{} scales to multiple references in a strictly plug-and-play manner, and that its training-free design can effectively integrate complementary in-context signals without dataset- or shot-specific adaptation.

\begin{table}[t]
\centering
\small
\setlength{\tabcolsep}{5.1pt}
\renewcommand{\arraystretch}{1.00}
\caption{\textbf{Backbone analysis for \ours{}.} We report mIoU (in \%, $\uparrow$) on COCO-20$^i$ of \ours{} with different backbone features. DINOv3 features lead to the best segmentation accuracy.}

\vspace{-4pt}
\begin{tabularx}{\columnwidth}{@{}ccccc@{}}
\toprule
\textbf{DINOv3} & \textbf{DINOv2} & \textbf{Franca} & \textbf{Perception Enc.} & \textbf{Stable Diff.}\\
\midrule
\bfseries \num{57.6} & \num{45.1} & \num{39.2} & \num{48.4} & \num{33.2} \\
\bottomrule
\end{tabularx}
\vspace{-0.3em}
\label{tab:backboneablation}
\end{table}

\subsection{Dense representations matter}

A central motivation of our work is that earlier ICS methods had to
\emph{compensate} for the limited spatial structure of existing VFM features,
either by training segmentation decoders \cite{Meng:2024:SEGiC}, fine-tuning
diffusion models \cite{Zhu:2024:Unleashing}, or coupling DINOv2 with SAM for
mask generation \cite{Zhang:2024:GF-SAM, Liu:2023:Matcher}. These
components were introduced because prior VFMs did not simultaneously provide \emph{(i)} reliable semantic correspondence across images and \emph{(ii)} sufficiently dense, part-aware structure within a single image. We validate this argument directly by replacing DINOv3 with other VFMs. In particular, we use features from DINOv2~\cite{Oquab:2023:Dinov2}, Franca~\cite{Venkataramanan:2025:Franca}, Perception Encoder~\cite{Bolya:2025:PE}, and Stable Diffusion 2.1~\cite{Rombach:2022:SD}. We apply \ours{} without any algorithmic changes, and tune hyperparameters for each backbone (\cf \cref{tab:backboneablation}). Segmentation accuracy of all other VFMs drops significantly on COCO-20$^i$ over DINOv3. These results support a key observation: DINOv3 self-supervised representations jointly exhibit strong semantic structure and spatially localized, dense features, which together are sufficient for ICS to arise from a single frozen backbone without decoders, fine-tuning, or external models.

\begin{table}[t]
\centering
\small
\setlength{\tabcolsep}{16pt}
\caption{\textbf{Comparison with SAM3.} We compare \ours{} to Segment Anything 3 (SAM 3) on COCO-20$^i$ using mIoU (in \%, $\uparrow$). We evaluate two adaptations of SAM 3: \emph{(i)} a \emph{video-style} formulation, where reference and target are treated as consecutive frames and the mask is propagated, and \emph{(ii)} an \emph{image concatenation} strategy, where images are combined horizontally and prompted.}
\vspace{-4pt}
\begin{tabularx}{\linewidth}{@{}Xc@{}}
\toprule
\textbf{Method} & \textbf{mIoU} \\
\midrule
\multicolumn{2}{@{}l@{}}{\textbf{Training free}: \textit{Mask-supervised pre-training}}\\
\;\;Segment Anything 3 (video-style propagation) & 26.7 \\
\;\;Segment Anything 3 (image concatenation) & 52.9 \\
\midrule
\multicolumn{2}{@{}l@{}}{\textbf{Training free}: \textit{Unsupervised pre-training}}\\
\;\;\textbf{INSID3} \emph{(ours)} & \bfseries 57.6 \\
\bottomrule
\end{tabularx}
\label{tab:sam3_comparison}
\end{table}

\subsection{Comparison with SAM 3}
\label{subsec:sam3}
We further compare \ours{} with the recent Segment Anything 3 (SAM 3) model~\cite{Carion:2025:SAM3}, a foundation model for promptable segmentation that supports multiple input modalities, including points, masks, and text prompts. Among its capabilities, SAM 3 enables \emph{visual prompting}, where a mask provided on an object can be used to segment other instances of the same concept \emph{within the same image}. In this experiment, we investigate whether such visual prompting can generalize across \emph{different} images, \ie from a reference to a target, to solve in-context segmentation. Since SAM 3 is not natively designed for this setting, we consider two adaptations. First, following prior work~\cite{Cuttano:2025:Sansa}, we adopt a \emph{video-style} formulation, where the reference and target images are treated as consecutive frames and the mask is propagated from the reference (first frame) to the target. Second, we propose an \emph{image concatenation} strategy, where the reference and target are stacked horizontally into a single image and the mask is provided on the reference region. Results on COCO-20$^i$ are reported in \cref{tab:sam3_comparison}. We observe that the image concatenation strategy significantly outperforms the video-style propagation, achieving \SI{52.9}{\percent} mIoU compared to a lower accuracy of \SI{26.7}{\percent}, respectively. Notably, \ours{} attains \SI{57.6}{\percent} mIoU, outperforming SAM3 by \num{4.7} percentage points despite not relying on any segmentation supervision.

\begin{table}[t]
\centering
\small
\setlength{\tabcolsep}{1pt}
\caption{\textbf{Absence of the reference concept.} Percentage of correct empty-mask predictions (\%, $\uparrow$) when the target image does not contain the prompted object (COCO, 4\,000 pairs). \textit{n.a.}: methods that always return a non-empty mask (\ie, \SI{0}{\%} in practice).}
\vspace{-4pt}
\label{tab:absence}
\begin{tabularx}{\linewidth}{@{}Xc}
\toprule
\textbf{Method} & \textbf{Correct empty predictions (in \%)}\ $\uparrow$ \\
\midrule
\multicolumn{2}{@{}l@{}}{\textbf{Task-specific fine-tuning}: \textit{Semantic + mask supervision}} \\
\;\;SegGPT \cite{Wang:2023:SegGPT} & 79 \\
\;\;SegIC \cite{Meng:2024:SEGiC} & 36 \\
\midrule
\multicolumn{2}{@{}l@{}}{\textbf{Training free}: \textit{Mask-supervised pre-training}} \\
\;\;PerSAM \cite{Zhang:2023:PerSAM} & \textit{n.a.} \\
\;\;Matcher \cite{Liu:2023:Matcher} & \textit{n.a.} \\
\;\;GF-SAM \cite{Zhang:2024:GF-SAM} & \textit{n.a.} \\
\midrule
\multicolumn{2}{@{}l@{}}{\textbf{Training free}: \textit{Unsupervised pre-training}} \\
\;\;\textbf{INSID3} \emph{(ours)} & \textbf{85} \\
\bottomrule
\end{tabularx}
\vspace{-0.5em}
\end{table}

\subsection{Corner case: Empty mask}
The standard in-context segmentation formulation assumes that the reference concept is present in the target image. However, a practical use case arises when the target image does not contain the object specified by the reference prompt, in which case the desired output is an empty mask. \ours{} can naturally accommodate this case by also subjecting the seed cluster to the aggregation criterion (\cf Eq.\ \ref{eq: seed_cluster_prot}). When the concept is absent, cross-image similarity scores remain uniformly low, and no cluster satisfies the selection criterion, resulting in an empty prediction.

To evaluate this setting, we construct 4\,000 reference--target pairs from COCO (80 classes), where the target image does not contain the reference concept. \Cref{tab:absence} reports the percentage of correct empty predictions. \ours{} correctly predicts empty masks in 85\% of cases. Methods based on SAM \cite{Liu:2023:Matcher, Zhang:2024:GF-SAM, Zhang:2023:PerSAM} cannot produce empty outputs by design and therefore always return a non-empty mask. Compared to supervised approaches, \ours{} outperforms SegIC \cite{Meng:2024:SEGiC} and SegGPT \cite{Wang:2023:SegGPT} despite requiring no task-specific training.

\definecolor{softheader}{RGB}{210,220,230}

\begin{table}[t]
\centering
\small
\setlength{\tabcolsep}{5.3pt}

\caption{\textbf{Computational analysis} of \ours{} \vs previous methods in milliseconds (ms, $\downarrow$). Inference time is measured for a single in-context example (\ie, example and target image). Runtime is measured on a single RTX 4090.}
\vspace{-0.4em}

\begin{tabularx}{\linewidth}{@{}X X c l@{}}
\toprule
\textbf{Method} & \textbf{Backbone} & \textbf{Resolution} & \textbf{Runtime} $\downarrow$ \\
\midrule

\multicolumn{4}{@{}l@{}}{\textbf{Task-specific fine-tuning}: \textit{Semantic + mask supervision}} \\
\;\;SegGPT \cite{Wang:2023:SegGPT} & {ViT} & 640\textsuperscript{2} & \hphantom{9}\,\SI{110}{ms} \\
\;\;SegIC \cite{Meng:2024:SEGiC}  & {DINOv2} & 896\textsuperscript{2} & \hphantom{9}\,\SI{301}{ms} \\
\midrule
\multicolumn{4}{@{}l@{}}{\textbf{Training free}: \textit{Mask-supervised pre-training}} \\
\;\;PerSAM \cite{Zhang:2023:PerSAM} & {SAM} & 1024\textsuperscript{2} & \hphantom{9}\,\SI{504}{ms} \\
\;\;Matcher \cite{Liu:2023:Matcher} & {DINOv2 + SAM} & 1024\textsuperscript{2} & \SI{9000}{ms} \\
\;\;GF-SAM \cite{Zhang:2024:GF-SAM} & {DINOv2 + SAM} & 1024\textsuperscript{2} & \SI{1030}{ms} \\
\midrule

\multicolumn{4}{@{}l@{}}{\textbf{Training free}: \textit{Unsupervised pre-training}} \\
\;\;\textbf{\ours{}} \emph{(ours)} & {DINOv3} & 1024\textsuperscript{2} & \bfseries \;\;\,\SI{302}{ms} \\

\bottomrule
\end{tabularx}

\vspace{-0.3em}
\label{tab:computational}
\end{table}

\begin{table}[t]
\centering
\small
\renewcommand{\arraystretch}{1.0}
\setlength{\tabcolsep}{2pt}
\caption{\textbf{Detailed computational analysis of \ours{}.} We report inference time of each component in milliseconds (ms, $\downarrow$). Inference time is measured for a single in-context example. Runtime measured on a single RTX 4090.}
\vspace{-0.4em}
\begin{tabularx}{\linewidth}{@{}Xr}
\toprule
\textbf{Component} & \textbf{Runtime} $\downarrow$ \\
\midrule
Encoder forward $+$ debiasing & \SI{78}{ms} \\
Compute similarity $+$ cluster scoring & \SI{3}{ms} \\
Clustering & 166 ms \\
CRF refinement & 55 ms \\
\midrule
\textbf{Total inference time} & \bfseries \SI{302}{ms} \\
\bottomrule
\end{tabularx}
\vspace{-0.3em}
\label{tab:computationaldetailed}
\end{table}
\begin{table}[ht]
    \centering
    \setlength{\tabcolsep}{1pt}
    \small
    \caption{\textbf{Computational analysis for different resolutions.} We report inference time of \ours{} for different resolution in milliseconds (ms, $\downarrow$). Inference time is measured for a single in-context example. Runtime measured on a single RTX 3090.}
    \vspace{-0.4em}
    \begin{tabularx}{\linewidth}{@{}X X X X X X X@{}}
        \toprule
        \multicolumn{7}{c}{\textit{Low resolution} $\leftarrow$ \textbf{Runtime of INSID3} $\rightarrow$ \textit{High resolution}} \\
        \midrule
        \bfseries\SI{512}{px} & \bfseries\SI{720}{px} & \bfseries\SI{896}{px} & \bfseries\SI{1024}{px} & \bfseries\SI{1440}{px} & \bfseries\SI{1600}{px} & \bfseries\SI{1760}{px} \\
        \hspace{4.5pt}\SI{93}{ms} & \SI{180}{ms} & \SI{230}{ms} & \hspace{5.5pt}\SI{302}{ms} & \hspace{5.5pt}\SI{689}{ms} & \hspace{5.5pt}\SI{930}{ms} & \SI{1102}{ms} \\
        \bottomrule
    \end{tabularx}%
    \label{tab:computationalresolution}
    \vspace{-0.3em}
\end{table}

\section{Computational Cost of \ours{}}
\label{sec:computational}

\Cref{tab:computational} reports the inference speed of \ours{} compared to training-free and fined-tuned baselines. \ours{} provides a significantly faster inference runtime than existing training-free approaches. This efficiency follows directly from our \emph{single-backbone design}. In contrast, SAM-based pipelines separate the problem into two stages: DINOv2 is used to
compute semantic correspondences, and SAM is prompted to produce masks. Because these two models
operate in different feature spaces, semantic matching and mask generation are decoupled and
must be coordinated through additional steps (\eg, point selection, prompt engineering, mask
filtering, and scoring). This multi-stage interaction results in significant inference overhead.

In contrast, \ours{} performs both semantic alignment and mask generation directly in the
DINOv3 feature space. 
On a single RTX~4090, \ours{} runs at \SI{302}{\ms}, compared to \SI{1030}{\ms} for GF-SAM, the most competitive baseline \wrt downstream accuracy.
The detailed runtime breakdown (\cf \cref{tab:computationaldetailed}) shows that the dominant cost of \ours{} is agglomerative clustering (\SI{166}{\ms}), with the
DINOv3 forward pass and debiasing accounting for \SI{78}{\ms}. Similarity computation and cluster
scoring are negligible (\SI{3}{\ms}). We use a GPU-accelerated dense CRF~\cite{Huu:2021:CRF} for refinement (\SI{55}{\ms}). Although agglomerative clustering has quadratic complexity in the number of tokens, \ours{} nevertheless scales well in practice to higher image resolutions (\cf \Cref{tab:computationalresolution}). Notably, \ours{} at \num{1760}$\times$\num{1760} remains comparable in runtime to GF-SAM at \num{1024}$\times$\num{1024}. Overall, \ours{} remains computationally efficient by operating entirely within a single feature space, without relying on auxiliary segmentation models such as SAM. We emphasize that all reported runtimes are measured in 32-bit precision and averaged over 100 examples. Additional speedups could be obtained through standard optimizations such as half-precision inference.

\section{Qualitative Examples}
\label{sec:qualitatives}
\begin{figure*}
        \centering
        \includegraphics[width=0.98\linewidth]{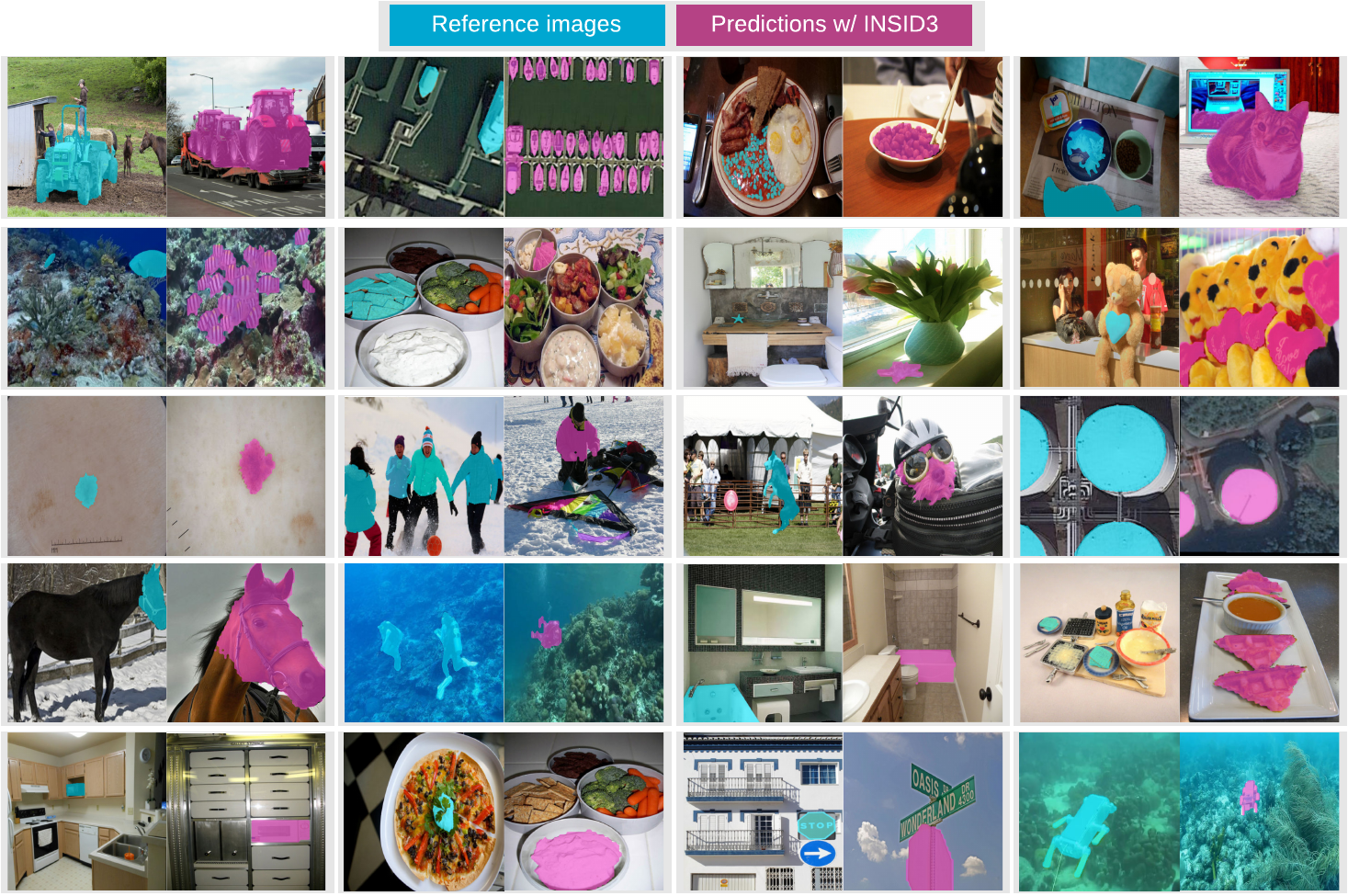}
    \vspace{-0.5em}
    \caption{\textbf{Qualitatives results with \ours{}}. Each pair shows the reference image with its annotated region \emph{(\textcolor{myblue}{light blue mask})} and the predicted mask on the target \emph{(\textcolor{mypurple}{purple})}. \ours{} handles a wide range of semantic granularities, from full objects to fine-grained parts, and generalizes robustly across diverse domains, including aerial, marine, medical, and everyday scenes.}
    \label{fig:qualitatives_supp}
    \vspace{-0.5em}
\end{figure*}
\Cref{fig:qualitatives_supp} presents additional qualitative results across a wide range of scenarios.
Each example shows the reference image with its annotated region \emph{(light blue)} alongside the predicted mask on the target image \emph{(purple)}.
Across a wide range of visual concepts, spanning different semantic granularities (from full objects to fine-grained parts) and diverse visual domains (aerial imagery, marine scenes, medical scans, and everyday images), \ours{} consistently produces coherent, accurate masks. It reliably identifies the correct concept even in the presence of distractor objects, similar classes, or large appearance changes, illustrating the consistency and versatility of \ours{}.

\section{Limitations and Future Work}
\label{sec:limitations}
\ours{} demonstrates that strong in-context segmentation capabilities can emerge directly from frozen self-supervised representations, without task-specific training or architectural modifications. While this simple formulation already achieves competitive results across diverse domains and granularities, some limitations remain. First, \ours{} currently handles one target concept at a time, requiring separate reference prompts when multiple concepts are present in the target image. Extending the method to jointly reason about multiple concepts within a single inference pass would be an interesting direction for future work. Second, \ours{} relies on \textit{masks} to define the target concept. In contrast, models such as Segment Anything can be prompted using lighter spatial annotations such as points or bounding boxes. These forms of annotation are typically cheaper and faster to collect, which could facilitate practical deployment and interactive use; incorporating lighter prompt modalities would therefore broaden the applicability of the approach. A related future direction concerns instance-level reasoning. While our focus is \textit{semantic} in-context segmentation, one could instead provide an example object in the reference image and ask the model to segment the corresponding instances in the target image separately, rather than merging them into a single semantic mask. In our observations, clusters belonging to the same instance often exhibit stronger mutual similarity than clusters from different instances. Exploiting these affinities to iteratively recover multiple instances, for example by expanding a seed with its most similar clusters and then re-seeding on the remaining regions, is an interesting direction for future work. Finally, \ours{} relies on the semantic structure encoded in frozen self-supervised features of DINOv3. Although our debiasing improves cross-image matching, the quality of the final segmentation still depends on the representational properties of the underlying backbone, suggesting that future advances in self-supervised representations could further strengthen this paradigm.

\end{document}